%% file: main.tex
\documentclass[10pt,twocolumn,letterpaper]{article}

\usepackage[pagenumbers]{cvpr} %

\input{preamble}



\definecolor{cvprblue}{rgb}{0.21,0.49,0.74}
\usepackage[pagebackref,breaklinks,colorlinks,allcolors=cvprblue]{hyperref}

\def\confName{CVPR}
\def\confYear{2027}

\title{\name: Semantic 3D Proxies for Controllable Agentic 3D Modeling}

\author{
    Shuyuan Zhang \quad Changjian Li \\
    The University of Edinburgh
}

\begin{document}
\maketitle

\begin{strip}
\centering
\includegraphics[width=\textwidth]{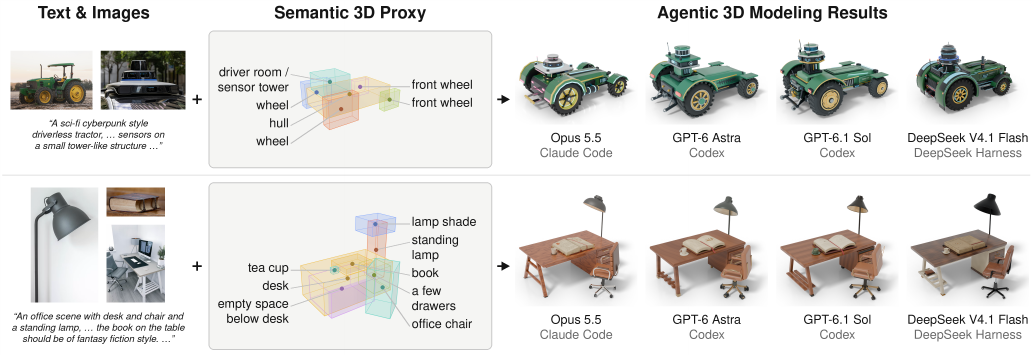}
\captionsetup{hypcap=false}
\captionof{figure}{\textbf{Semantic 3D proxies bring 3D-native control to agentic 3D modeling.}
In addition to text and reference images, semantic 3D proxies, \eg, labeled bounding boxes, provide an intuitive interface for users and a structured representation directly consumable by coding agents, enabling effective control over 3D modeling results across different agent backends.
}
\captionsetup{hypcap=true}
\label{fig:teaser}%
\end{strip}

\input{sec/0_abstract}
\input{sec/1_intro}
\input{sec/2_related_work}
\input{sec/3_method}
\input{sec/4_evaluation}
\input{sec/5_conclusion}

{
    \small
    \bibliographystyle{ieeenat_fullname}
    \bibliography{main}
}

\input{sec/X_suppl}

\end{document}

%% file: preamble.tex
\usepackage{cuted} %
\usepackage{caption} %

\usepackage{xcolor}
\usepackage{multirow}
\usepackage{colortbl}
\usepackage{xspace}
\usepackage{graphicx}

\usepackage{amssymb}%

\newcommand{\name}{\emph{SemProxy}\xspace}
\newcommand{\uiname}{\emph{ProxyStudio}\xspace}

\newcommand{\sBase}{T+I\xspace}
\newcommand{\sGT}{T+I+P\xspace}
\newcommand{\sNoisy}{T+I+P\textsubscript{noised}\xspace}
\newcommand{\sDrafted}{T+I+P\textsubscript{bootstrapped}\xspace}
\newcommand{\sBaseFull}{Text + Images\xspace}
\newcommand{\sGTFull}{w/ Semantic 3D Proxy\xspace}
\newcommand{\sNoisyFull}{w/ Noised Proxy\xspace}
\newcommand{\sDraftedFull}{w/ Bootstrapped Proxy\xspace}
\newcommand{\fittab}[1]{\resizebox{\ifdim\width>\linewidth\linewidth\else\width\fi}{!}{#1}}
\newcommand{\sRenderOnly}{Render-only\xspace}
\newcommand{\sTextTemplate}{Text-template\xspace}
\newcommand{\sTextLLM}{Text-LLM\xspace}

%% file: sec/0_abstract.tex
\begin{abstract}
Coding agents are increasingly capable of generating 3D models, yet enabling users to effectively control the modeling process remains challenging.
Existing workflows rely primarily on text prompts and reference images, which often struggle to convey precise spatial intentions and require iterative corrections.
We introduce \name, a lightweight intermediate representation that bridges intuitive human interaction and agentic 3D modeling.
Instantiated as labeled 3D bounding boxes, semantic 3D proxies allow users to directly specify object parts and spatial arrangements while leaving geometric details to coding agents.
These proxies can be serialized into structured text and rendered as images, making them directly accessible to existing coding agents without additional training or modifications.
We further develop \uiname, a user interface that integrates text, image references, and interactive 3D proxy editing into agentic modeling workflows.
We demonstrate the versatility of \name across object- and scene-level modeling tasks with different coding agents, and evaluate its effectiveness in supporting controllable human-agent collaborative 3D modeling through a user study.

\end{abstract}

%% file: sec/1_intro.tex
\section{Introduction}
\label{sec:intro}

\begin{figure}[t]
\centering
\includegraphics[width=\linewidth]{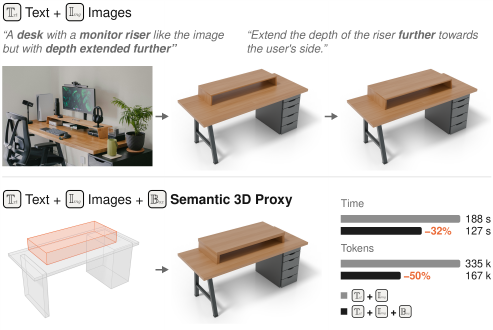}
\caption{
\textbf{Semantic 3D proxies make spatial intent explicit.} Text and images alone may require iterative corrections to communicate precise 3D structures (top). Semantic 3D proxies express these structures directly through labeled bounding boxes, enabling more efficient and controllable modeling (bottom). `Time' is the agent's total run time over all turns, and `Tokens' is input and output tokens over the whole conversation.
}
\label{fig:ideation}
\end{figure}

Large language models (LLMs) are increasingly capable of creating 3D models by generating code through geometry and modeling APIs, either in dedicated pipelines~\cite{sun20253dgpt,hu2024scenecraft} or through general-purpose coding agents such as Claude Code and Codex~\cite{gao20263dcodebench,liu20263dharnessbench}.
However, enabling users to effectively control these agents remains a challenge.
Existing workflows primarily rely on text prompts and reference images, which can convey the overall appearance of a desired model but often struggle to communicate precise 3D modeling intentions, such as part dimensions, spatial arrangements, and relative positions.
Describing such details in words is tedious, while inferring them from images can be ambiguous, often requiring repeated interactions to achieve the intended result.
As illustrated in \cref{fig:ideation} (top), even a simple modification to the depth of a desk's monitor riser requires an additional prompting round to clarify the desired geometry.

We argue that controllable agentic 3D modeling calls for a more direct way to communicate 3D intent. Rather than relying entirely on text and images to describe spatial structures indirectly, users should be able to specify them directly in 3D. However, requiring users to construct detailed 3D models would defeat the purpose of agentic modeling. An ideal representation should therefore be lightweight and intuitive, allowing users to express only the spatial and semantic information they care about, while leaving the remaining modeling details to the agent.

Achieving this requires a representation that bridges human interaction and agent understanding. On the one hand, it should be \emph{3D-native}, allowing users to intuitively specify and manipulate spatial structures directly in 3D. On the other hand, it should be \emph{agent-native}, encoding these structures in forms that existing coding agents can readily interpret without specialized encoders or additional training. While text and images are naturally accessible to these agents, they do not directly provide editable 3D structures. Conversely, dense 3D representations such as meshes and point clouds are cumbersome for users to author and are not readily interpretable by off-the-shelf coding agents~\cite{wang2024llamamesh, dai2025meshcoder}.

To bridge this gap, we propose \emph{semantic 3D proxies} (\name), a lightweight representation that combines semantic labels with simple 3D geometry. We instantiate this representation using labeled 3D bounding boxes, which users can easily create, position,
and resize to specify the parts and spatial arrangements of a desired model. These proxies can be serialized into structured text and rendered as images, allowing existing coding agents to directly interpret the user-specified 3D structures. Importantly, the proxies need not describe the complete model: users can specify only the parts they wish to control, at their preferred level of detail. As shown in \cref{fig:ideation} (bottom), a simple bounding box specifying only the rough shape of the intended monitor riser enables the agent to produce the desired geometry without additional prompting.

We evaluate \name across proprietary and open-source coding agents on both object- and
scene-level modeling tasks. With semantic 3D proxies, all agents produce models that
better match the intended geometry at a lower cost, and participants in our preference
survey preferred them over models built from text and images in 67.5\,\% of collected responses.
Our contributions are:
\begin{itemize}
    \item We introduce \name, a human-friendly and agent-native representation for controllable 3D modeling. Instantiated as labeled 3D bounding boxes, semantic 3D proxies enable intuitive spatial control across a diverse range of modeling tasks, without requiring agent fine-tuning or modifications to existing modeling APIs.

    \item We develop \uiname, a user interface that integrates text, image references, and interactive 3D proxy editing with coding agents. We further conduct a user study to evaluate its effectiveness in supporting controllable human-agent collaborative 3D modeling.
\end{itemize}

%% file: sec/2_related_work.tex
\section{Related Work}
\label{sec:rw}

\begin{figure*}[t]
\centering
\includegraphics[width=\textwidth]{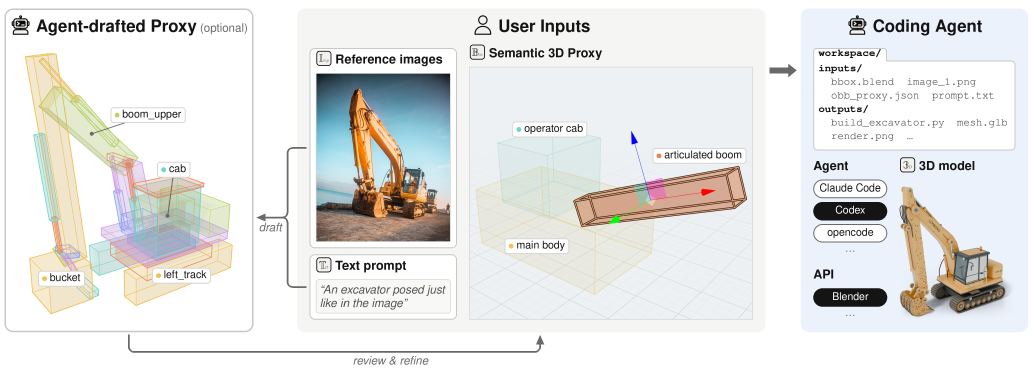}
\caption{\textbf{Overview of the \name workflow.}
Users specify their modeling intent through text, reference images, and semantic 3D proxies edited in \uiname (middle). Optionally, an agent can draft an initial proxy from text and images for users to refine (left).
An off-the-shelf coding agent then consumes these inputs to construct the 3D model through a modeling API, here Blender (right).
}
\label{fig:method}
\end{figure*}

\paragraph{Agentic 3D modeling.}
A growing body of work casts 3D modeling as program synthesis: an LLM writes code against a modeling API, runs it and refines the result from renders.
Training-free systems orchestrate pretrained LLMs and VLMs to set procedural parameters~\cite{sun20253dgpt}, compose retrieved assets~\cite{hu2024scenecraft}, search over program edits~\cite{huang2024blenderalchemy}, split object modeling into stages~\cite{lu2025ll3m,zhang2025shapecraft}, recover programs from images~\cite{yin2026viga,he2026thinking} or generate CAD models with visual self-verification~\cite{alrashedy2025cadcodeverify,mallis2025cadassistant,badagabettu2024query2cad,fan2026caddesigner}.
Trained models translate text, images or point clouds into Blender or CAD code~\cite{du2024blenderllm,dai2025meshcoder,rukhovich2025cadrecode,kolodiazhnyi2026cadrille,xu2024cadmllm,guan2025cadcoder,khan2024text2cad} or mesh tokens~\cite{wang2024llamamesh,fang2025meshllm,ye2025shapellmomni}, trading generality for accuracy within a domain.
Others first commit to an intermediate structure, such as assembly graphs~\cite{lin2026procedura}, DSL programs~\cite{wang2026adsl}, part and joint hierarchies~\cite{gao2026lam,zhou2026articraft}, scene programs~\cite{zhang2025scenelanguage,tam2025scenemotifcoder,liu2025agentic,sun20253dgeneralist} or scene layouts~\cite{feng2023layoutgpt,zhou2024gala3d,yang2024holodeck,sun2025layoutvlm}.
Most recently, off-the-shelf coding agents such as Claude Code and Codex model through general-purpose APIs, as evaluated by 3DCodeBench~\cite{gao20263dcodebench} and 3DHarnessBench~\cite{liu20263dharnessbench} alongside other program-based benchmarks~\cite{gu2025blendergym,yang2026p3dbench}.

\paragraph{Inputs to agentic systems.}
These systems are conditioned almost exclusively on text and images.
Where 3D input is used, it is a dense reconstruction target, such as the point clouds of trained code generators~\cite{dai2025meshcoder,rukhovich2025cadrecode,kolodiazhnyi2026cadrille} or the full target scene in 3DHarnessBench~\cite{liu20263dharnessbench}.
Bounding boxes appear as plans that the LLM writes from text, as in the per-part volumes of ShapeCraft~\cite{zhang2025shapecraft} and generated scene layouts~\cite{feng2023layoutgpt,zhou2024gala3d,yang2024holodeck}, and users correct them through further dialogue~\cite{lu2025ll3m,fan2026caddesigner}.
In \name, users author the labeled boxes, which condition a general-purpose coding agent without fine-tuning it or extending the modeling API.

\paragraph{3D proxies in generation.}
Cuboids and other primitives have long served as shape abstractions~\cite{tulsiani2017learning,ye2025primitiveanything,mo2019partnet,mo2019structurenet}, and ShapeAssembly~\cite{jones2020shapeassembly} writes programs of cuboid parts, but there boxes are the output, not a condition.
Many generative models take coarse proxies as control: primitive assemblies and guidance shapes~\cite{dong2024coin3d,sella2024spicee}, part handles~\cite{hertz2022spaghetti,koo2023salad}, user-placed part boxes~\cite{slim2026compose,pun2026multicube,koo2026boxsplitgen}, constraint meshes~\cite{dihlmann2026arbor}, and boxes as plans, part prompts or loose image conditions~\cite{yang2025omnipart,ding2025fullpart,yan2025xpart,bhat2024loosecontrol}.
Some frameworks need no training: SpaceControl~\cite{fedele2026spacecontrol} guides a pretrained generator with primitives at test time, HandMade~\cite{huang2026handmade} renders part-labeled VR sketches as guidance for existing generators, and in Prox-E~\cite{sella2026proxe}, a pretrained VLM reads primitives as text and rendered views and edits them to guide a 3D generator.
In scenes, models generate box layouts~\cite{paschalidou2021atiss,tang2024diffuscene}, and user-placed proxies condition synthesis~\cite{cohenbar2023setthescene,yang2024scenecraft}, such as the \emph{semantic proxy room} of ControlRoom3D~\cite{schult2024controlroom3d}, whose term we partly borrow.
3D language models are also trained to read and write boxes~\cite{cho2025cubellm,avetisyan2024scenescript,mao2025spatiallm,wang2025partxmllm,wang2026artllm}.
In all of these, the proxy guides a 3D generative model or is read by a language model trained for it. \name gives it to unmodified coding agents as text and rendered views, and the agents build the model through a modeling API.

%% file: sec/3_method.tex
\section{Method}
\label{sec:method}

\name conditions a coding agent on a \emph{semantic 3D proxy} alongside text and images
(\cref{fig:method}). We first define the proxy in general (\cref{sec:proxy}), then
instantiate it with labeled bounding boxes (\cref{sec:boxes}), describe how a proxy reaches
an agent through a workspace (\cref{sec:workspace}), and how the agent can draft the proxy
for the user (\cref{sec:draft}).

\subsection{Semantic 3D Proxies}
\label{sec:proxy}
A semantic 3D proxy is a set of elements $\mathcal{P}=\{p_k\}_{k=1}^{K}$, each pairing
\emph{semantic} information with \emph{spatial} information. The semantic part is text: a
label naming what the element stands for, optionally with a free-form description. The
spatial part says where the element is and how large it is, and may take any geometric
form, \eg, a box or another shape primitive. We place two
requirements on the proxy.
(i) \emph{3D-native}: the spatial part is explicit 3D geometry with exact values. It states
3D information directly, and the user places and drags it in 3D instead of describing it in
words, which is intuitive and user-friendly.
(ii) \emph{Agent-native}: the proxy takes the form of text and/or images, the domains that
language models are trained on, so no encoder or fine-tuning is needed. Every element can be
written as structured text without loss, so the agent reads exact values instead of
estimating them from pixels.
To fully leverage the information encoded in the proxy, it can also be drawn as a 3D scene that the agent
renders and compares with the reference images and its own output.
A proxy need not be complete: its elements can be parts of an object or objects in a scene,
at whatever granularity the user cares about, and the text and images specify or imply
the rest.

\subsection{Labeled Bounding Boxes}
\label{sec:boxes}
We instantiate semantic 3D proxies with labeled oriented bounding boxes, the simplest form we
know of that meets both requirements. Each element is
$p_k=(\ell_k,\mathbf{c}_k,\mathbf{s}_k,\mathbf{R}_k)$: a semantic component $\ell_k$ that
carries text, a center $\mathbf{c}_k\in\mathbb{R}^3$, extents
$\mathbf{s}_k\in\mathbb{R}^3$ along the box's own axes, and an orientation
$\mathbf{R}_k\in SO(3)$. We serialize the boxes as a JSON list of these fields
(\texttt{obb\_proxy.json}) and store them as a 3D scene (\texttt{bbox.blend}) with one cuboid per box, named by its
label. Boxes are familiar from any 3D editor and quick to place, move
and resize.
For extension, a new form of proxy would change only these two files, and the rest of the method would be unchanged.

\begin{table*}[!t]
\centering
\small
\caption{\textbf{Main results.} Each agent's modeling performance and cost given the caption and the reference images alone (\emph{\sBaseFull}), with GT bounding boxes added as the semantic 3D proxy (\emph{\sGTFull}, \cref{subsec:abl}), with a proxy that an agent bootstrapped from the caption and images (\emph{\sDraftedFull}, \cref{subsec:comp}), or with the GT proxy perturbed by 10\% noise (\emph{\sNoisyFull}, \cref{subsec:discussion}). Chamfer distance to the GT mesh: geometric mean, $\times 10^{3}$; DINOv3 and Uni3D similarity to the GT mesh: mean; 20 samples per level. Cost per trial is averaged over all samples, and the bootstrapped proxy's cost counts both the box-drafting and the modeling sessions. \textbf{Bold}: best per agent; \underline{underlined}: second best.}
\fittab{\input{tab/tab_overview_core}}
\label{tab:main}
\end{table*}

\subsection{Workspace}
\label{sec:workspace}
Each modeling task runs in a workspace folder with two subfolders. The folder \texttt{inputs/} holds
what the user provides: the text instruction (\texttt{prompt.txt}); reference images (\texttt{image\_N.png}), each optionally paired with a text note
(\texttt{image\_N.txt}) saying what to take from that image, \eg, following the
fantasy style of the book or borrowing the sensor elements of the driverless car
in \cref{fig:teaser}; and the proxy as \texttt{obb\_proxy.json} and
\texttt{bbox.blend}.
The folder \texttt{outputs/} stores all of the agent's work: scripts, renders,
and the final model. The agent starts in the workspace with a short task prompt that points it to \texttt{inputs/} (prompts in the supplementary material).
The prompt asks for a model that meets the joint
expectation of the text, the images and the boxes, with the boxes as approximate references
rather than exact targets. The agent saves its work, including the final 3D model, to
\texttt{outputs/}. To refine the
result, the user edits the inputs between turns, and the same agent session continues,
told what changed.

The method fixes only this input end. Which coding agent and harness run in the workspace,
and how they build geometry, are left open: any agent that can read files, and any modeling
tool or API it can drive, would serve. We use Blender's Python API as the geometry backend
throughout, and evaluate three off-the-shelf coding agents in \cref{sec:experiments}.

\paragraph{Interface.}
\uiname (\cref{fig:method}, middle) is our interface to the workspace. It writes
\texttt{inputs/} from a 3D box editor, a text field and image cards, and launches a
coding agent, \eg, Codex, Claude Code, OpenCode or the DeepSeek Harness, with any model it
supports. It shows \texttt{outputs/} in
the same viewport, with the model over its boxes (full screenshot in the supplementary material).

\subsection{Drafting the Proxy}
\label{sec:draft}
Laying out every box by hand is not required. The user can instead ask the agent for a
draft: the agent reads the text instruction and images and writes boxes to \texttt{outputs/},
checking them against the references through renders of the boxes. The interface shows
the draft next to any boxes the user has already placed, and the user reviews, edits and
accepts it. Accepted boxes move to \texttt{inputs/} and are treated exactly like boxes the
user placed, so modeling proceeds as in \cref{sec:workspace}.

%% file: tab/tab_overview_core.tex
\setlength{\tabcolsep}{3pt}
\begin{tabular}{@{}llrrrrrrrrrrrr@{}}
\toprule
 & & \multicolumn{3}{c}{Easy objects} & \multicolumn{3}{c}{Hard objects} & \multicolumn{3}{c}{Scenes} & \multicolumn{3}{c}{Cost per trial} \\
\cmidrule(lr){3-5}\cmidrule(lr){6-8}\cmidrule(lr){9-11}\cmidrule(l){12-14}
Agent & Setting & \multicolumn{1}{c}{Chamfer$\downarrow$} & \multicolumn{1}{c}{DINOv3$\uparrow$} & \multicolumn{1}{c}{Uni3D$\uparrow$} & \multicolumn{1}{c}{Chamfer$\downarrow$} & \multicolumn{1}{c}{DINOv3$\uparrow$} & \multicolumn{1}{c}{Uni3D$\uparrow$} & \multicolumn{1}{c}{Chamfer$\downarrow$} & \multicolumn{1}{c}{DINOv3$\uparrow$} & \multicolumn{1}{c}{Uni3D$\uparrow$} & \multicolumn{1}{c}{\shortstack{Time$\downarrow$\\(min)}} & \multicolumn{1}{c}{\shortstack{Input$\downarrow$\\(M)}} & \multicolumn{1}{c}{\shortstack{Output$\downarrow$\\(k)}} \\
\midrule
\multirow{4}{*}{Codex} & \textit{\sBaseFull} & 5.6 & 0.886 & 0.841 & 5.6 & 0.854 & 0.836 & 7.8 & 0.809 & 0.829 & 4.2 & \textbf{0.27} & 7.2 \\
 & \textbf{\textit{\sGTFull}} & \textbf{4.1} & \textbf{0.901} & \textbf{0.859} & \textbf{2.5} & \textbf{0.880} & \textbf{0.866} & \textbf{3.2} & \textbf{0.820} & \textbf{0.835} & \underline{4.1} & \underline{0.29} & \textbf{6.7} \\
 & \textit{w/ Bootstrapped Proxy} & \underline{4.4} & 0.883 & \underline{0.853} & 5.6 & 0.853 & 0.839 & 6.0 & 0.808 & \underline{0.834} & 4.7 & 0.55 & 9.3 \\
 & \textit{\sNoisyFull} & 4.5 & \underline{0.889} & 0.847 & \underline{3.1} & \underline{0.870} & \underline{0.849} & \underline{4.3} & \underline{0.811} & 0.830 & \textbf{3.0} & 0.31 & \underline{7.0} \\
\midrule
\multirow{4}{*}{Claude Code} & \textit{\sBaseFull} & 5.5 & \textbf{0.851} & \underline{0.835} & 5.6 & 0.825 & 0.835 & 9.7 & 0.782 & 0.807 & \underline{2.5} & \textbf{0.31} & \underline{11.9} \\
 & \textbf{\textit{\sGTFull}} & \textbf{3.3} & \underline{0.843} & \textbf{0.848} & \textbf{2.6} & \textbf{0.856} & \underline{0.842} & \textbf{3.9} & 0.779 & \underline{0.808} & \textbf{2.3} & \underline{0.35} & \textbf{11.2} \\
 & \textit{w/ Bootstrapped Proxy} & 5.2 & 0.820 & 0.819 & 6.4 & 0.816 & 0.827 & 8.1 & \textbf{0.786} & 0.805 & 3.6 & 0.51 & 15.0 \\
 & \textit{\sNoisyFull} & \underline{3.4} & 0.830 & 0.824 & \underline{3.3} & \underline{0.854} & \textbf{0.850} & \underline{4.4} & \underline{0.785} & \textbf{0.810} & 2.6 & 0.39 & 12.0 \\
\midrule
\multirow{4}{*}{DeepSeek} & \textit{\sBaseFull} & 6.7 & 0.830 & \underline{0.815} & 6.5 & 0.813 & 0.797 & 10.5 & 0.765 & 0.807 & 36.6 & 30.0 & 243 \\
 & \textbf{\textit{\sGTFull}} & \underline{4.9} & \textbf{0.851} & 0.804 & \textbf{3.8} & \textbf{0.843} & \textbf{0.824} & \textbf{3.8} & \textbf{0.789} & \textbf{0.826} & \textbf{34.6} & \textbf{26.8} & \textbf{217} \\
 & \textit{w/ Bootstrapped Proxy} & 6.5 & 0.823 & 0.813 & 6.0 & 0.818 & 0.795 & 7.1 & \underline{0.775} & \underline{0.819} & 43.4 & 29.4 & 289 \\
 & \textit{\sNoisyFull} & \textbf{4.2} & \underline{0.837} & \textbf{0.828} & \underline{4.4} & \underline{0.831} & \underline{0.810} & \underline{4.7} & 0.771 & \underline{0.819} & \underline{35.7} & \underline{28.5} & \underline{222} \\
\bottomrule
\end{tabular}

%% file: sec/4_evaluation.tex
\section{Experiments}
\label{sec:experiments}

\noindent\textbf{Implementation.}
We run three off-the-shelf coding agents with no modifications and at their default levels of reasoning effort: Codex (GPT-6.1 Sol, medium), Claude Code (Claude Opus 5.5, medium) and the open-source DeepSeek Harness (DeepSeek V4.1 Flash, high).
Each agent works in the workspace of \cref{sec:workspace} with Blender's Python API, for up to 120\,min per session, and we run one trial per sample, setting and agent (full prompts in the supplementary material).

\noindent\textbf{Test sets.}
We curated two test sets for our evaluation.
Every sample has a caption, a ground-truth (GT) mesh and labeled bounding boxes (GT proxy).
For objects, we sample 40 shapes from PartVerse~\cite{dong2025copart} with Cap3D captions~\cite{luo2023cap3d} and one box per part: 20 \emph{easy} objects with 3--6 parts and 20 \emph{hard} objects with 10--19 parts.
For scenes, we sample 20 indoor rooms from MIDI's test split~\cite{huang2025midi} of 3D-FRONT~\cite{fu20213dfront}, each with 3--7 objects whose boxes carry their 3D-FUTURE category~\cite{fu20213dfuture} and 3D-GRAND description~\cite{yang20253dgrand}.
Three renders of the GT mesh serve as reference images (\cref{fig:qualitative}, left).
The construction process is detailed in the supplementary.

\noindent\textbf{Metrics.}
We evaluate geometry quality with the Chamfer distance~\cite{fan2017point}, and visual quality with two similarity scores: the image similarity of DINOv3~\cite{simeoni2025dinov3} between standardized renders of the agent's mesh and of the GT mesh, and the shape similarity of Uni3D~\cite{zhou2024uni3d} between the two meshes.
The detailed metric equations, further geometric metrics and the agreement of each model with its input text and images are reported in the supplementary.

\subsection{Ablation Study}
\label{subsec:abl}

To evaluate the effectiveness of the semantic 3D proxy, we compare two settings:
\begin{itemize}
\item \emph{\sBaseFull} (\sBase): the agent models from the caption of the sample as the text prompt and the three renders of its GT mesh as reference images.
\item \emph{\sGTFull} (\sGT): the agent additionally gets the labeled bounding boxes of the sample as the semantic 3D proxy (\cref{sec:boxes}).
\end{itemize}
Note that we use GT proxy in place of user-crafted proxy for a fair and controlled comparison, since the latter tend to vary from person to person.

\begin{figure}[!t]
\centering
\includegraphics[width=\linewidth]{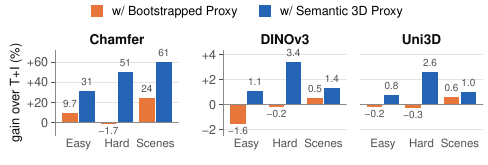}
\caption{\textbf{Gain of the semantic 3D proxy by difficulty.} Relative improvement over \sBase of \emph{\sGTFull} (blue) and of \emph{\sDraftedFull} (orange, \cref{subsec:comp}) on each metric, averaged over the three agents, on easy objects, hard objects and scenes. The bootstrapped proxy keeps a small part of the gain or falls below \sBase.}
\label{fig:gain}
\end{figure}

Quantitatively, \cref{tab:main} shows that the agents achieve both better geometry accuracy and better visual quality when modeling with the semantic 3D proxy, while spending less time and fewer output tokens.
This holds in the majority of cases, across proprietary and open-source coding agents and on both object and scene modeling.
On average, the Chamfer distance improves by 47\%, DINOv3 similarity by 2\% and Uni3D similarity by 1.5\%. The count of input tokens slightly grows naturally for including more information, and in practice they are usually priced much lower than output tokens when it comes to monetary costs.
Notably, the open-source DeepSeek agent achieves similar modeling performance and statistical patterns, though with much longer time and higher token consumption.
The gain is larger on harder tasks (\cref{fig:gain}, blue).
On Chamfer distance, it grows from easy objects to hard objects and to scenes.
On the similarity scores, it is largest on hard objects and saturates on scenes, where it is still larger than on easy objects.

\begin{figure*}[!t]
\centering
\includegraphics[width=\textwidth]{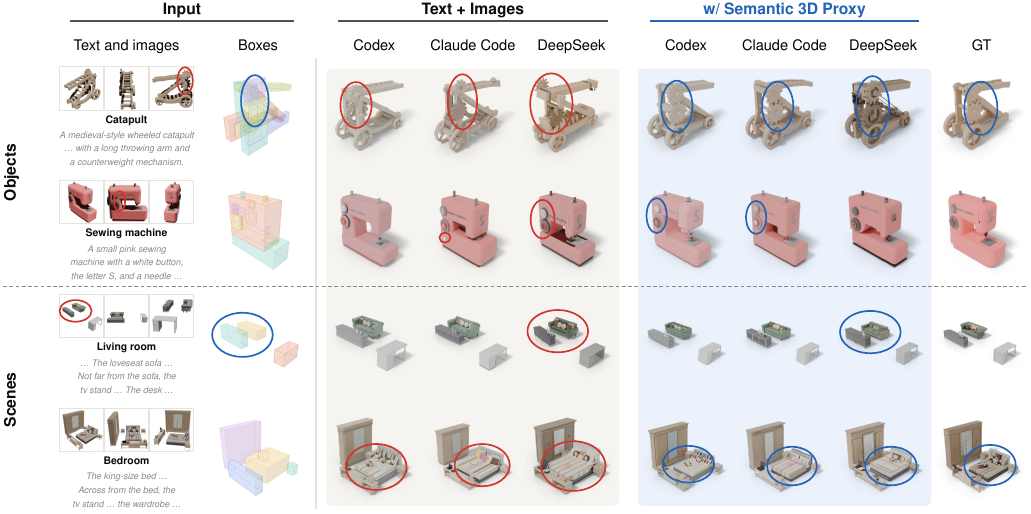}
\caption{\textbf{Qualitative comparison.} Each row shows the inputs (the caption, abridged at ``\dots'', the three reference images and the GT boxes given as the proxy), the three agents' models in \emph{\sBaseFull} and in \emph{\sGTFull}, and the GT mesh. Ellipses mark the parts discussed in the text: red in the reference images and in the models built from text and images, blue in the proxy, in the models built with it and in the GT.
}
\label{fig:qualitative}
\end{figure*}

Qualitatively, the \sBase baseline suffers from occlusion in the images and from the difficulty of conveying exact sizes and layouts in natural language (\cref{fig:qualitative}).
In the catapult example, the gear set on the side is visible but heavily occluded by its own support structure, so each agent builds a different gear mechanism in \sBase; the proxy preserves this information, and all agents build the right gear structure.
For the sewing machine, one reference image shows the elements on its side clearly, yet the agents in \sBase still arrange the buttons differently and vary the overall proportions.
For scenes, the main difference is the accuracy of the layout, such as the relative positions of the sofa and the TV stand in the living room, or the size of the bed in the bedroom.

\begin{figure}[t]
\centering
\includegraphics[width=\linewidth]{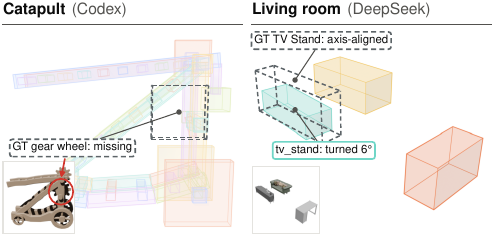}
\caption{\textbf{Bootstrapped proxies} for the catapult and the living room of \cref{fig:qualitative}; dashed: GT boxes. Insets: reference images, with the occluded gear wheel circled in red.
}
\label{fig:bootstrap_proxy}
\end{figure}

\begin{figure*}[t]
\centering
\includegraphics[width=\textwidth]{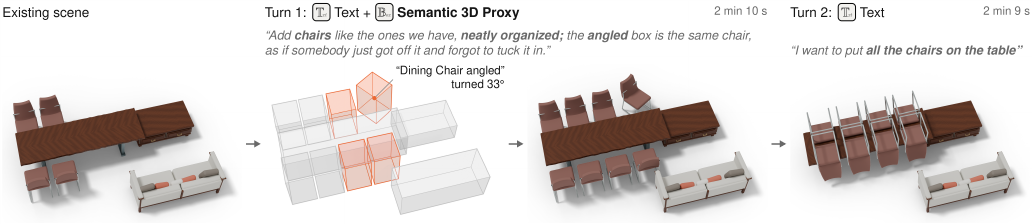}
\caption{\textbf{Example editing trace} from the user study (T3). A participant continues the agent's model of a
room. They copy a dining chair's box four times, place the copies and turn one of them, and ask in text for matching
chairs, the turned one left pulled out from the table. A second request, in text alone, moves all the chairs onto the
table. The instructions are shortened; times are each turn's wall time (Codex, GPT-6.1 Sol, medium effort).
}
\label{fig:editing_traces}
\end{figure*}

\subsection{Comparison}
\label{subsec:comp}

Coding agents can also draft the proxy for the user from the text and image inputs (\cref{sec:draft}), but how much of the underlying 3D information can they recover, or can they relieve the user from manipulating the boxes at all when text and images already say what the user wants?
We therefore evaluate a third setting, \emph{\sDraftedFull} (\sDrafted): one agent session first drafts the proxy from the inputs of \sBase, and a second session takes this proxy as if the user had provided it and models as in \sGT.

In \cref{tab:main}, \sDrafted falls behind \sGT on almost every level and metric. In \cref{fig:gain} (orange), we show that its overall performance is closer to \sBase than to \sGT, where it either keeps a limited gain comparing to what GT proxy brings, or performs even worse than the \sBase baseline.

We argue that this is because a bootstrapped proxy does not carry the 3D information that GT proxy or a user-issued box does.
The setting is effectively \sBase with an additional instruction to plan first with a coarse proxy.
Where 3D information is missing from the inputs, due to the same occlusion and ambiguity as in \sBase, the agent that drafts the boxes cannot reliably recover it either.
Although the bootstrapping instruction directs the agent to read the layout from the inputs first, which brings the limited gains, the whole pipeline still operates on the same amount of 3D information as \sBase.
When the agent recovers the wrong 3D information, error amplification effect in this process will then negatively impact the agent's modeling capabilities.

Our observation on the actual agent traces supports this argument, and we show two representative types of errors that we see in the bootstrapped proxies in \cref{fig:bootstrap_proxy}.
In the catapult's case, the heavily occluded gear wheel on the side is missing, which is the same mistake as in the results of \sBase as shown earlier in \cref{fig:qualitative}.
In the living room's case, the agent misjudges the orientation of the TV stand since the perspective of the image makes it look like it's slightly rotated - the same mistake of DeepSeek in the baseline \sBase setting as shown in \cref{fig:qualitative}.

\subsection{User Study}
\label{subsec:user_study}

The experiments above use GT boxes in place of a user's boxes to keep the comparison controlled.
We complement them with two studies with real human users: whether they prefer the models built with the semantic 3D proxy over the baseline, and how they experience agentic 3D modeling with labeled bounding boxes as additional control.

\paragraph{Preference.}
In an online survey, 32 participants compared the models that an agent built for the same sample in \sBase and in \sGT.
Over 535 responses, they preferred the model built with the proxy in 67.5\%.
The detailed procedure and correlation analysis between human preference and automated metrics can be found in the supplementary material.

\paragraph{Modeling with the proxy.}
Seven participants, with 3D modeling experience from none to extensive, modeled objects and edited a scene in \uiname, with the exact user study available in the supplemental material.
Generally, they felt in control of the modeling results with the help of semantic 3D proxy, and found the final results satisfactory. Although bootstrapped proxy performs badly in our quantitative experiments, participants accepted the first drafts or only issued simple editing instructions to correct the agent-drafted boxes.
It is also common that human users tend to use text prompts for general instructions, reference images for matching visual details, and semantic 3D proxy for precise spatial control.
The example editing trace visualized in \cref{fig:editing_traces} shows representative use of text, to describe high-level editing intentions about adding or moving the chairs, and of semantic 3D proxy, to specifically tell the agent which chair to rotate and to what extent should it be different from the other chairs.

\subsection{Discussions}
\label{subsec:discussion}

\paragraph{Robustness to rough proxies.}
We have been using GT labeled bounding boxes as perfect proxies, and they show the performance upper-bound as if these proxies perfectly reflect user's modeling intentions.
In practice, we do not want to put extra burden on the user's side by asking them to always issue perfect proxies, so we explore how the system react to noises and inaccuracies in the semantic 3D proxy.
To this end, we introduce the \emph{\sNoisyFull} (\sNoisy) setting, which perturbs each GT bounding box by up to 10\% of its extent in position and scale, and turns the box by up to $5^\circ$ about a random axis.
Evaluated on the full test set, \cref{tab:main} shows that the agents' performance under this setting stays close to \sGT with the GT proxy, and often comes second best, beating the \sDrafted setting.
This shows some level of robustness, and that even a rough proxy still carries 3D information better than what the agent can recover from text and images alone.

\begin{figure}[!t]
\centering
\includegraphics[width=0.97\linewidth]{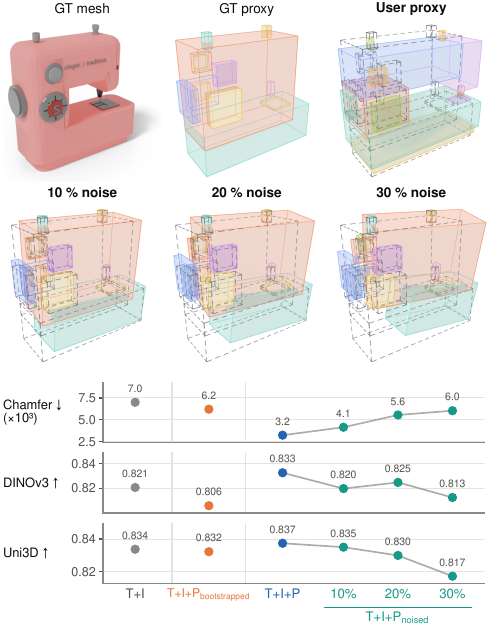}
\caption{\textbf{Effect of proxy noise.} Top: an object's GT mesh and GT proxy, a user proxy, and the proxy at 10--30\% noise; the user proxy and the noised proxies (color) are drawn over the GT proxy (dashed). Bottom: Chamfer distance (geometric mean, $\times 10^{3}$) and DINOv3 and Uni3D similarity (mean) to the GT mesh on the 15-sample subset, averaged over the three agents.
}
\label{fig:noise}
\end{figure}

To further investigate the robustness of semantic 3D proxies, we increase the noise level to 20\% and 30\%, with angle changes of up to $10^\circ$ and $15^\circ$ respectively, and test on a smaller 15-sample subset of 5 easy objects, 5 hard objects, and 5 scenes sampled from our full test set.
The upper part of \cref{fig:noise} shows the noised proxies alongside GT and user proxy. At 30\% noise, the structure is heavily lost, and it is unlikely that the agent can obtain any useful 3D information from the noised proxy other than a named part list. In comparison, although an user-issued proxy might still differ from GT, it preserves the overall structure well and its parts also stay coherent.
We also observe an overall performance drop as the noise level increases in the lower part of \cref{fig:noise}.
Notably, at 10\% or 20\% noise, agent performance still stays close to \sGT or the other baselines, but at 30\% noise, both DINOv3 and Uni3D similarity drop below the \sBase baseline. This shows that at the 30\% level, the agent cannot effectively infer the true 3D information behind the noises, and thus any conflicting information the agent perceives from the inputs will start to negatively impact its 3D modeling capabilities.

\begin{table}[!t]
\centering
\caption{\textbf{Proxy encoding.} Chamfer distance and similarity to the GT mesh on the 15-sample subset, by encoding of the GT proxy, averaged over the three agents. Chamfer distance: geometric mean, $\times 10^{3}$; DINOv3 and Uni3D similarity: mean. \textbf{Bold}: best; \underline{underlined}: second best. Per-agent results are in the supplementary material.}
\label{tab:encoding}
\footnotesize
\fittab{\input{tab/tab_encoding}}
\end{table}

\paragraph{Proxy encodings.}
The experiments above show that modeling benefits from the 3D information that the proxy carries.
Does the gain hold if the same occlusion and loss of information are addressed with more traditional and less structured representations?
We thus replace the structured JSON and the Blender scene in the \sGT setting with alternative representations and evaluate them on the same 15-sample subset:
\begin{itemize}
    \item \sRenderOnly: labeled bounding boxes as colored and tagged renderings from 4 different viewing angles;
    \item \sTextTemplate: labeled bounding boxes described in natural language, generated from a template;
    \item \sTextLLM: also as natural language, but written by an LLM, given the full set of inputs as the \sGT setting.
\end{itemize}

As shown in \cref{tab:encoding}, no alternative encoding reaches the performance of the \sGT setting, with \sRenderOnly consistently beating other proxy encodings. This shows that images are better than pure text when it comes to preserving 3D information, and that half-transparent box renders partially solve the occlusion problem from the \sBase setting.

\begin{figure}[t]
\centering
\includegraphics[width=\linewidth]{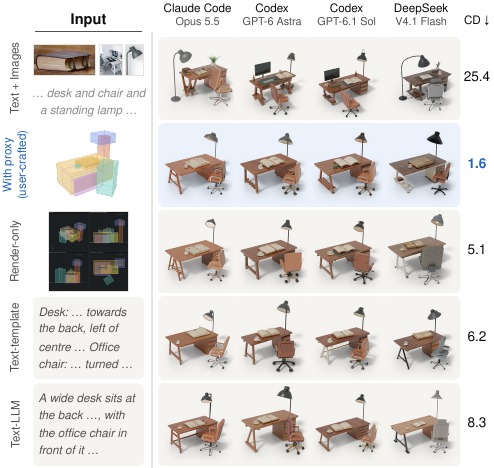}
\caption{\textbf{Controllability.} The office scene of \cref{fig:teaser} modeled by four agents under five settings; \emph{Input} shows what each setting adds to the text prompt and the reference images. With user-crafted proxy, all agents build the same layout. A render or a text description of the same layout leaves sizes and placement to each agent, and \emph{Text + Images} leaves the whole layout open. CD: Chamfer distance between two agents' delivered models, averaged over pairs of agents, $\times 10^{3}$.
}
\label{fig:controllability}
\end{figure}

\paragraph{Controllability.}
We further demonstrate the controllability proxies can achieve in terms of delivering consistent modeling results. In \cref{fig:controllability}, we compare the same office scene shown in the teaser image, built from user-issued proxies, with other baseline settings.
Qualitatively, the results modeled with semantic 3D proxy follow the specified layout and general user intentions better, while other proxy encodings deliver varying 3D models.
Quantitatively, semantic 3D proxy also achieves better average Chamfer distance between delivered 3D models, showing the consistency it can achieve across different coding agents.

%% file: tab/tab_encoding.tex
\begin{tabular}{@{}lrrr@{}}
\toprule
Setting & \multicolumn{1}{c}{Chamfer$\downarrow$} & \multicolumn{1}{c}{DINOv3$\uparrow$} & \multicolumn{1}{c}{Uni3D$\uparrow$} \\
\midrule
\textit{\sBaseFull} & 7.0 & 0.821 & 0.834 \\
\textbf{\textit{\sGTFull}} & \textbf{3.2} & \textbf{0.833} & \textbf{0.837} \\
\midrule
\textit{\sRenderOnly} & \underline{4.7} & \underline{0.825} & \underline{0.836} \\
\textit{\sTextTemplate} & 5.7 & 0.815 & 0.829 \\
\textit{\sTextLLM} & 5.9 & 0.813 & 0.835 \\
\bottomrule
\end{tabular}

%% file: sec/5_conclusion.tex
\section{Conclusion, Limitation, and Future Work}
\label{sec:conclusion}

We presented \name, which introduces semantic 3D proxies alongside text and images to communicate spatial intent more precisely to off-the-shelf coding agents, without requiring architectural changes or fine-tuning. By being both 3D-native and agent-native, these proxies enable more intuitive control over agentic 3D modeling.

Our evaluation has two main limitations. First, the quantitative experiments use ground-truth meshes and boxes in place of user-authored proxies, because no dataset offers user-placed 3D boxes at scale, and building one would take extensive crowdsourcing. Second, \name targets creative workflows that model novel assets, where no ground truth exists. In these workflows, as in \cref{fig:teaser}, users borrow elements from several sources, and the semantic 3D proxy gives them direct control over where and how each element enters the new model. We expect the proxy to help most in exactly these cases, yet they cannot be scored against a reference. We hope this work inspires richer spatial interfaces and more effective human-agent collaboration in 3D modeling and beyond. More discussions and typical failure cases can be found in the supplementary.

%% file: sec/X_suppl.tex
\clearpage
\appendix
\setcounter{table}{0}
\renewcommand{\thetable}{A\arabic{table}}
\setcounter{figure}{0}
\renewcommand{\thefigure}{A\arabic{figure}}
\setcounter{equation}{0}
\renewcommand{\theequation}{A\arabic{equation}}

\newenvironment{promptblock}{\par\vspace{3pt}\begingroup\footnotesize\ttfamily\raggedright
  \setlength{\parindent}{0pt}\setlength{\parskip}{3pt}\hyphenchar\font=-1 }{\par\endgroup\vspace{3pt}}

\newcommand{\mSecWorkspace}{\cref{sec:workspace}}
\newcommand{\mSecDraft}{\cref{sec:draft}}
\newcommand{\mSecAbl}{\cref{subsec:abl}}
\newcommand{\mSecUser}{\cref{subsec:user_study}}
\newcommand{\mTabMain}{\cref{tab:main}}
\newcommand{\mTabEncoding}{\cref{tab:encoding}}
\newcommand{\mFigIdeation}{\cref{fig:ideation}}
\newcommand{\mFigGain}{\cref{fig:gain}}
\newcommand{\mFigTrace}{\cref{fig:editing_traces}}

\maketitlesupplementary

The appendices give the details of the experiments (\cref{sec:supp_setup}),
additional results (\cref{sec:supp_results}), the details of the two user studies (\cref{sec:supp_user})
and further discussion (\cref{sec:supp_discussion}).
The settings are named as in the main paper: \emph{\sBaseFull} (\sBase), \emph{\sGTFull} (\sGT),
\emph{\sDraftedFull} (\sDrafted) and \emph{\sNoisyFull} (\sNoisy).

\section{Experiment Details}
\label{sec:supp_setup}

\subsection{Data Curation}
\label{sec:supp_data}

\paragraph{Objects.}
We join the 12{,}030 part-segmented objects of PartVerse~\cite{dong2025copart} with their Cap3D
captions~\cite{luo2023cap3d}. We then apply automatic filtering to keep individual objects with
3--20 parts with matching caption and segmentation. \emph{Easy} objects have 3--6
parts and a caption of at most 18 words, with their parts being mostly axis-aligned and primitive-like;
\emph{hard} objects have at least 10 parts and a caption of at least 12 words. From the two pools we draw
candidates over ten object types, and manually review their caption, renders and segmentation.

\paragraph{Scenes.}
The rooms and their object meshes come from MIDI's processed test split of
3D-FRONT~\cite{huang2025midi,fu20213dfront}: 1{,}000 rooms, each with one textured mesh per
object. We join each room with its scene description and object descriptions from
3D-GRAND~\cite{yang20253dgrand}, with the object categories
of 3D-FUTURE~\cite{fu20213dfuture}, and with the placements of the original 3D-FRONT house files,
which give each object's yaw.
We sample candidates over room types, one room per house, and we review each for
correct category information and consistent scale in the descriptions
against renders of the actual models.

\paragraph{Reference images.}
Each GT mesh is rendered in Blender with fixed lighting from three fixed directions: a
three-quarter view, the front and the side.

\paragraph{Semantic 3D Proxy from GT bounding boxes.}
Object part boxes are fitted to each PartVerse part's faces: the axis-aligned box, else a box
turned about the vertical axis, else the minimum-volume oriented box, each kept only when it
is at least 10\,\% tighter. Every box was then reviewed by hand so that boxes follow the
object's axes unless the part is truly slanted (mostly tilted boxes on
round or organic parts reset to axis-aligned). Scene boxes are fitted to each object's mesh,
axis-aligned or turned about the vertical by the object's yaw from the 3D-FRONT house file
when it is more than $2^\circ$ off the room's axes.

\subsection{Metric Definitions}
\label{sec:supp_metrics}

We define the three scores of \mTabMain{} and how they are aggregated. Each score
compares the mesh $\hat{M}$ that an agent delivers with the GT mesh $M$ of the sample. The protocol
follows recent benchmarks of agentic 3D modeling~\cite{gao20263dcodebench,liu20263dharnessbench}.
Geometry is scored on points alone, and the two similarity scores see the colors of the meshes.

\paragraph{Point sets.}
From each mesh we sample $N=8192$ points uniformly by surface area, center them on their centroid
and scale them so that the farthest point lies on the unit sphere. $\hat{X}$ and $X$ denote the
point sets of $\hat{M}$ and $M$.

\paragraph{Chamfer distance.}
We use the symmetric squared Chamfer distance~\cite{fan2017point} between the two point sets:
\begin{equation}
\begin{split}
d_{\mathrm{CD}}(\hat{X},X) ={} & \frac{1}{|\hat{X}|}\sum_{\hat{x}\in\hat{X}}\min_{x\in X}\lVert\hat{x}-x\rVert_2^2 \\
 & + \frac{1}{|X|}\sum_{x\in X}\min_{\hat{x}\in\hat{X}}\lVert x-\hat{x}\rVert_2^2 .
\end{split}
\label{eq:chamfer}
\end{equation}
Text and images do not fix which way a model faces, so we turn the prediction about the vertical
axis by $\theta\in\Theta=\{0^\circ,90^\circ,180^\circ,270^\circ\}$ and keep the best turn:
\begin{equation}
\theta^\star=\operatorname*{arg\,min}_{\theta\in\Theta}\, d_{\mathrm{CD}}(R_\theta\hat{X},X),
\qquad
\mathrm{CD}=d_{\mathrm{CD}}(R_{\theta^\star}\hat{X},X),
\label{eq:turn}
\end{equation}
where $R_\theta$ is the rotation by $\theta$ about the vertical axis. We report
$\mathrm{CD}\times10^{3}$. Every metric below that needs an aligned prediction uses the same
turn $\theta^\star$.

\paragraph{DINOv3 image similarity.}
We render $\hat{M}$ and $M$ in color with one rig: four views at azimuths $45^\circ$,
$135^\circ$, $225^\circ$ and $315^\circ$, $25^\circ$ above the horizon, at $512\times512$ pixels.
Each mesh keeps its own materials and is framed on its bounding sphere, and the lights move with
the camera, so every view of every mesh is lit alike. Let $\hat{I}_i$ and $I_j$ be the views of
$\hat{M}$ and $M$, and $\phi(\cdot)$ the unit-length class-token embedding of DINOv3
ViT-L/16~\cite{simeoni2025dinov3}. Since the prediction may be turned, the views are matched
one-to-one by the best assignment instead of by index:
\begin{equation}
s_{\mathrm{DINO}}=\max_{\pi\in\Pi_4}\ \frac{1}{4}\sum_{i=1}^{4}\phi(\hat{I}_i)^{\top}\phi(I_{\pi(i)}),
\label{eq:dino}
\end{equation}
where $\Pi_4$ is the set of permutations of the four views.

\paragraph{Uni3D shape similarity.}
Uni3D~\cite{zhou2024uni3d} embeds colored point clouds. Each point of $\hat{X}$ and $X$ takes
the base color of its mesh at that point. With $\psi(\cdot)$ the unit-length embedding of
Uni3D-g,
\begin{equation}
s_{\mathrm{Uni3D}}=\psi(R_{\theta^\star}\hat{X})^{\top}\psi(X).
\label{eq:uni3d}
\end{equation}
Uni3D was trained on single objects, so its scores on scenes are only indicative.

\paragraph{Aggregation.}
\mTabMain{} reports each score per level, over the 20 samples of the level: the
geometric mean for the Chamfer distance, whose distribution is skewed, and the arithmetic mean
for the two similarity scores. Every trial delivered a loadable mesh, so every mean is over all
samples. The relative gain of a setting $s$ over \sBase is computed from these level means, per
agent:
\begin{equation}
g_{\mathrm{CD}}=1-\frac{\mathrm{CD}_{s}}{\mathrm{CD}_{\text{\sBase}}},
\qquad
g_{\mathrm{sim}}=\frac{s_{s}}{s_{\text{\sBase}}}-1 ,
\label{eq:gain}
\end{equation}
so that a positive gain is an improvement for every score. \mFigGain{} shows the gain per level, averaged over the three agents, and the overall gains quoted
in \mSecAbl{} also average over the three levels.

\paragraph{Cost.}
The cost per trial in \mTabMain{} is the wall-clock time of the agent's session,
its input tokens, which include cached ones, and its output tokens, each averaged over all 60
samples. For \sDrafted, the drafting and the modeling sessions are summed.

\subsection{Prompts}
\label{sec:supp_prompts}

All agents receive the same prompts, given verbatim below. These are generic instructions and the same
for all modeling tasks, since the prompts point the agents to the files in \texttt{inputs/}, instead
of dynamically loading different text descriptions or images for each task.

\paragraph{Modeling with a proxy.}
\sGT, \sNoisy and the modeling session of \sDrafted share one prompt. Its last paragraph asks
the agent for renders of its own model from the viewpoints of the reference images, on which
the agreement with the inputs is scored (\cref{sec:supp_agreement}):
\begin{promptblock}
Task: write python scripts using the bpy library to create a 3D model described by the text,
images, and bounding boxes under inputs/.

The text description is in inputs/prompt.txt, and you can find user-supplied images in inputs/
too. Each image is optionally accompanied by a short description in a txt file with the same
name explaining how to interpret that image. The oriented bounding boxes are in
inputs/obb\_proxy.json and inputs/bbox.blend - treat them as approximate 3D references rather
than exact targets. Aim for a good-quality model that meets the joint expectation of the text
description, the reference images, and the box proxies.

Your workspace is limited to this directory containing inputs/ and outputs/ only. Put all your
work in outputs/, like scripts or any renderings, and save the final mesh as outputs/mesh.glb.
\textasciigrave bpy\textasciigrave{} is installed in a uv-managed venv, so use \textasciigrave
uv run python\textasciigrave{} to execute scripts that need it. Check for GPU availability
first if you want to use GPU to accelerate your rendering.

Additional task: when you finish modeling, also provide your own renderings of the model
matching the reference images visually (e.g. styles, viewing angles, etc.) and with
corresponding names (e.g. outputs/image\_1.png for inputs/image\_1.png).
\end{promptblock}

\paragraph{Modeling without a proxy.}
The prompt of \sBase differs only where the proxy is named. Its first paragraph ends
``\ldots described by the text and images under inputs/.'', and its second paragraph has no
sentence about the boxes and ends ``\ldots meets the joint expectation of the text description
and the reference images.''

\paragraph{Drafting.}
The first session of \sDrafted writes the proxy. Its workspace has the caption, the reference
images and a tool that renders the boxes:
\begin{promptblock}
Task: write a JSON file of labeled 3D bounding boxes laying out a 3D model described by the
text and images under inputs/.

The text description is in inputs/prompt.txt, and you can find user-supplied images in inputs/
too. Each image is optionally accompanied by a short description in a txt file with the same
name explaining how to interpret that image. Use one box per part or per object, at the level
of detail the text and images call for, and aim for a layout whose overall size, proportions
and arrangement meet the joint expectation of the text description and the reference images.

Save the boxes as outputs/bbox.json, in this shape:

\{"boxes": [\{"label": "seat", "center": [0, 0, 0.45], "dimensions": [0.5, 0.5, 0.08],
"rotation\_deg": [0, 0, 0]\}]\}

Right-handed with +Z up; the model's front faces -Y and it stands on the ground plane z = 0.
center is the box centre; dimensions are its full extents along its own X, Y and Z axes;
rotation\_deg is optional, XYZ Euler angles in degrees. Use short, distinct names as labels.

Your workspace is limited to this directory containing inputs/, outputs/ and tools/ only. Put
all your work in outputs/. \textasciigrave bpy\textasciigrave{} is installed in a uv-managed
venv, so use \textasciigrave uv run python\textasciigrave{} to execute scripts that need it. To
check your boxes, run \textasciigrave uv run python tools/render\_bboxes.py\textasciigrave: it
validates outputs/bbox.json, prints each box's extents and renders the boxes in perspective,
front, right and top views to outputs/renders/bbox\_views.png, which you can examine and
compare with the input conditions.
\end{promptblock}

\paragraph{Agent-specific note.}
Claude Code's prompts end with one more paragraph about its shell tool, which returns after
120\,s while the command keeps running: the note tells the agent to set a longer timeout for
builds and renders instead of starting them again. Codex and the DeepSeek Harness get no note.

\subsection{Interface}
\label{sec:supp_ui}

\begin{figure*}[t]
\centering
\includegraphics[width=\textwidth]{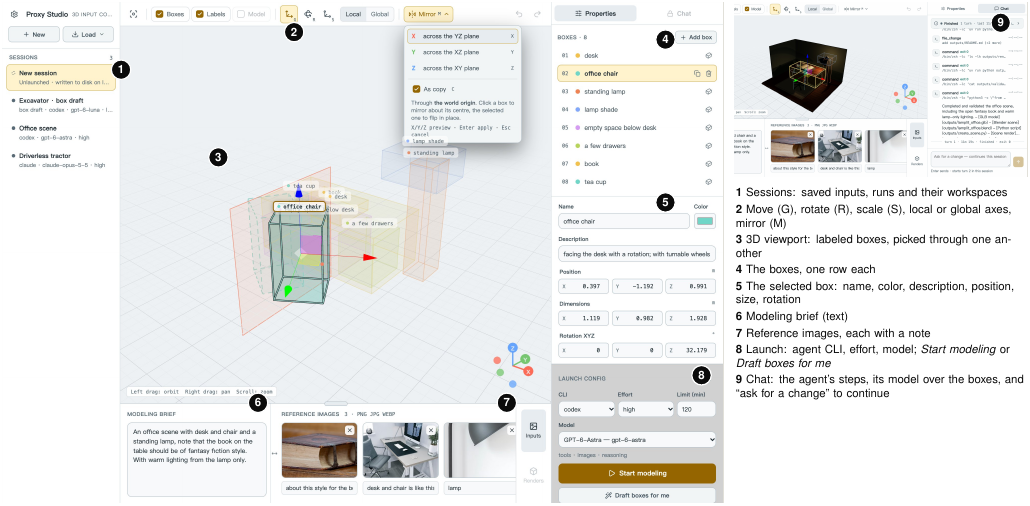}
\caption{\textbf{\uiname.} Left: composing the inputs of an office scene, with one box selected. Right: the session after the agent finished, with its model shown over the boxes and the conversation open for a follow-up turn. The numbers mark the parts of the interface.}
\label{fig:studio_ui}
\end{figure*}

\cref{fig:studio_ui} shows \uiname, our interface to the workspace of \mSecWorkspace{}.
The numbers below are those in the figure.

\paragraph{Composing the inputs.}
The proxy is edited in the 3D viewport (3), which draws every box with its label. Boxes are
semi-transparent and can be picked through one another.
The toolbar (2) offers the tools of a 3D editor with Blender-style hotkeys: move (G),
rotate (R) and scale (S), in the box's own axes or in the global axes, and a mirror tool (M)
that copies or flips boxes about a plane. Boxes can also be duplicated and deleted, and every
edit can be undone. The panel on the right lists the boxes (4) and shows the selected one (5):
its name, color and description, and its exact position, size and rotation, in meters and
degrees, which can be typed in. Below the viewport, the user writes the text prompt (6) and adds
reference images, each with an optional note that says what to take from it (7).

\paragraph{Launching an agent.}
The launch panel (8) selects the coding agent, its model and its reasoning effort.
\emph{Start modeling} writes the text, the images and the proxy to \texttt{inputs/} and starts
the agent in the workspace. \emph{Draft boxes for me} starts a drafting session instead
(\mSecDraft{}): the drafted boxes appear next to the boxes that the user has
already placed, and the user accepts them one by one or all together. The session list (1) keeps
every session with its inputs, its runs and its workspace, so that a session can be reopened and
continued.

\paragraph{Continuing a session.}
While the agent works, the boxes are locked and the chat (9) shows the agent's steps and its
renders; the user can stop the agent at any time. When the agent finishes, its model appears in
the same viewport over the boxes and can be switched on and off. To refine the result, the user
edits the inputs, types a request in the chat, or both: edits made between turns are sent with
the next message, and the same agent session continues, told what changed.

\section{Additional Results}
\label{sec:supp_results}

\cref{tab:all_metrics} gives every score of a setting averaged over the three agents and the three
levels: the scores of \mTabMain{}, two further geometry metrics
(\cref{sec:supp_geometry}) and the agreement of each model with its input text and images
(\cref{sec:supp_agreement}). \cref{sec:supp_encoding} gives the results of the proxy encodings
per agent.

\begin{table*}[t]
\centering
\small
\setlength{\tabcolsep}{5pt}
\fittab{\input{tab/tab_all_metrics_mean}}
\caption{\textbf{All metrics, by setting.} The scores of \mTabMain{}, two further geometry metrics and the agreement of each model with its input text and images, averaged over the three agents and the 60 samples. SigLIP~2 text--render scores the delivered model, and the two agent-render metrics score the renders that the agent made of its own model. Chamfer distance: geometric mean, $\times 10^{3}$. \textbf{Bold}: best; \underline{underlined}: second best.}
\label{tab:all_metrics}
\end{table*}

\subsection{Additional Geometry Metrics}
\label{sec:supp_geometry}

Besides the Chamfer distance, \cref{tab:all_metrics} reports two further geometry metrics. Both
use the point sets and the turn $\theta^\star$ of \cref{sec:supp_metrics}; we write
$\hat{X}^\star=R_{\theta^\star}\hat{X}$ and $d(p,Y)=\min_{y\in Y}\lVert p-y\rVert_2$.

\paragraph{F-score.}
The F-score~\cite{tatarchenko2019single} at a threshold $\tau$ is the harmonic mean of
precision, the share of predicted points within $\tau$ of the GT, and recall, the share of GT
points within $\tau$ of the prediction:
\begin{equation}
\begin{gathered}
P_\tau=\frac{1}{|\hat{X}^\star|}\sum_{\hat{x}\in\hat{X}^\star}\big[d(\hat{x},X)\le\tau\big],\\
R_\tau=\frac{1}{|X|}\sum_{x\in X}\big[d(x,\hat{X}^\star)\le\tau\big],\\
\mathrm{F}@\tau=\frac{2\,P_\tau R_\tau}{P_\tau+R_\tau}.
\end{gathered}
\label{eq:fscore}
\end{equation}
We report F@0.05: $\tau=0.05$ is 2.5\,\% of the diameter of the unit sphere.

\paragraph{Surface-voxel IoU.}
sv-IoU compares the cells that the two surfaces occupy. With $V(Y)$ the set of cells of a $64^3$
grid over $[-1,1]^3$ that hold at least one point of $Y$,
\begin{equation}
\text{sv-IoU}=\frac{|V(\hat{X}^\star)\cap V(X)|}{|V(\hat{X}^\star)\cup V(X)|}.
\label{eq:sviou}
\end{equation}
It counts surface cells and not the enclosed volume, because the GT meshes are not watertight.

\begin{figure*}[t]
\centering
\includegraphics[width=\textwidth]{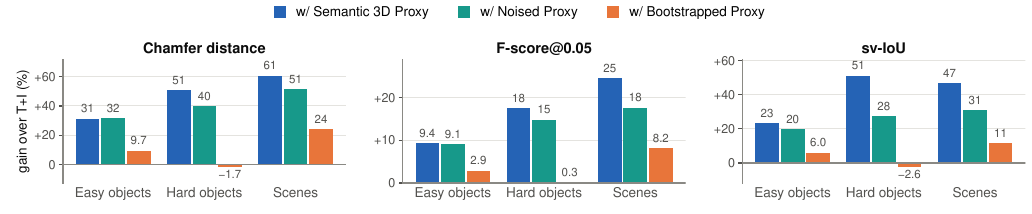}
\caption{\textbf{Geometry gains on all three metrics.} Relative improvement over \sBase in Chamfer distance, F-score and sv-IoU, averaged over the three agents, for the semantic 3D proxy (\sGT), the noised proxy (\sNoisy) and the bootstrapped proxy (\sDrafted). Each metric has its own scale.}
\label{fig:supp_geometry}
\end{figure*}

\paragraph{Results.}
The F-score and sv-IoU confirm what the Chamfer distance shows in the main paper
(\cref{tab:all_metrics,fig:supp_geometry}). With the semantic 3D proxy, all three metrics improve
for every agent at every level: averaged over the gains of the agents and levels, the Chamfer
distance drops by 47\,\%, the F-score rises by 17\,\% and sv-IoU by 40\,\%. Per sample, the model built
with the proxy is the better one for 148 of the 180 pairs of agent and sample in Chamfer distance,
for 143 in F-score and for 139 in sv-IoU. The three metrics also rank the settings alike. The gain
of the proxy is larger on hard objects and scenes than on easy objects, the noised proxy keeps
most of it, and the bootstrapped proxy brings a small gain on easy objects and scenes and none on
hard objects.

\subsection{Agreement with the Inputs}
\label{sec:supp_agreement}

The scores above compare a model with the GT mesh. The agreement metrics compare it with what
the agent was given: the caption $t$ and the three reference images $J_1,J_2,J_3$. All three are
cosine similarities between unit-length embeddings (\cref{tab:all_metrics}, last three columns).

\begin{figure}[t]
\centering
\includegraphics[width=\linewidth]{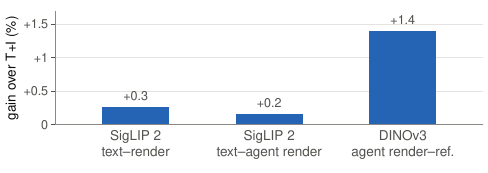}
\caption{\textbf{Agreement with the inputs.} Relative improvement of \sGT over \sBase on the three agreement metrics, averaged over the three agents and the three levels.}
\label{fig:supp_agreement}
\end{figure}

\paragraph{Agreement with the text.}
\emph{SigLIP~2 text--render} is the similarity of the caption to the four renders $\hat{I}_i$ of
\cref{eq:dino}, under the text and image encoders of SigLIP~2~\cite{tschannen2025siglip2},
averaged over the renders. \emph{SigLIP~2 text--agent render} is the same similarity to the
renders that the agent makes of its own model, described next. A caption that is longer than the
context window of the text encoder is split into sentences, and the scores of the sentences are
averaged; this applies to the scene captions.

\paragraph{Agreement with the images.}
The modeling prompt asks the agent to render its finished model so that the renders match the
reference images in style and viewing angle (\cref{sec:supp_prompts}). This is an inverse
graphics task: the agent reproduces each reference image with its own model, camera, lights and
materials. The renders $A_1,A_2,A_3$ show the model as the agent made it in Blender, and every
trial delivered all three. \emph{DINOv3 agent render--reference} is the similarity of each render
to the reference image that it was asked to match, $\frac{1}{3}\sum_{k}\phi(A_k)^{\top}\phi(J_k)$,
with the DINOv3 embedding $\phi$ of \cref{eq:dino}.

\paragraph{Results.}
The semantic 3D proxy also brings the models closer to their inputs
(\cref{tab:all_metrics,fig:supp_agreement}). The agents' renders match the reference images
better: DINOv3 agent render--reference rises from 0.793 to 0.804, by 1.4\,\%, close to the 2\,\%
that DINOv3 gains against the GT mesh in the main paper. The agreement with the text is kept: the
two SigLIP~2 scores change by at most 0.3\,\%. The proxy does not change how an agent approaches
the modeling task, and none of the three metrics looks at the boxes. The 3D information of the
proxy helps the agents reproduce the reference images more closely, beyond matching the boxes,
and at no cost to the text.

\subsection{Proxy Encodings: Per-Agent Results}
\label{sec:supp_encoding}

\begin{figure*}[t]
\centering
\includegraphics[width=\textwidth]{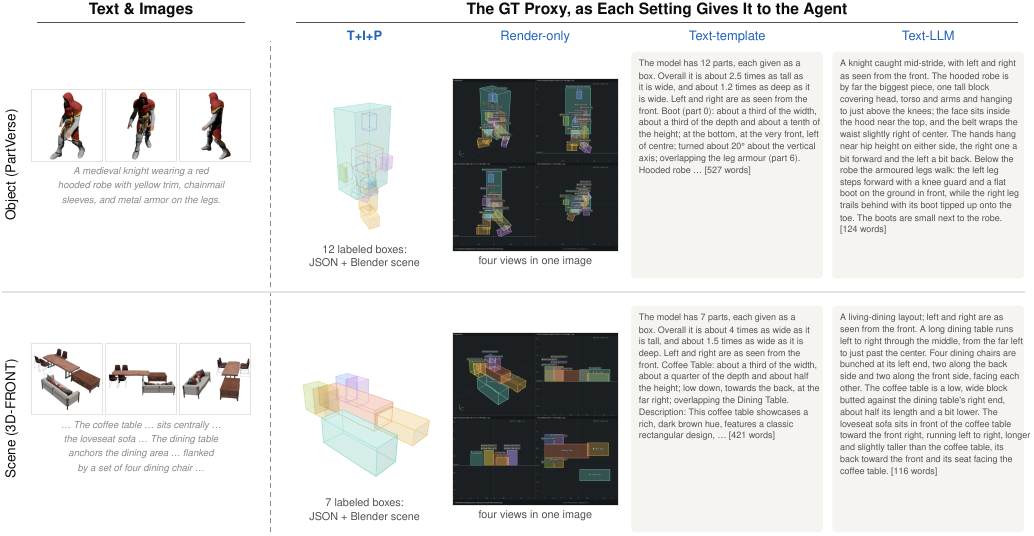}
\caption{\textbf{The GT proxy in each encoding}, for an object and a scene. Every setting also gets the caption and the three reference renders (left). \sGT: the labeled boxes as JSON and a Blender scene (drawn here). \sRenderOnly: one image of the boxes from four views, without box files. \sTextTemplate: generated sentences (the first 88 words shown). \sTextLLM: the LLM's paragraph, in full.}
\label{fig:encoding_inputs}
\end{figure*}

\begin{figure*}[t]
\centering
\includegraphics[width=\textwidth]{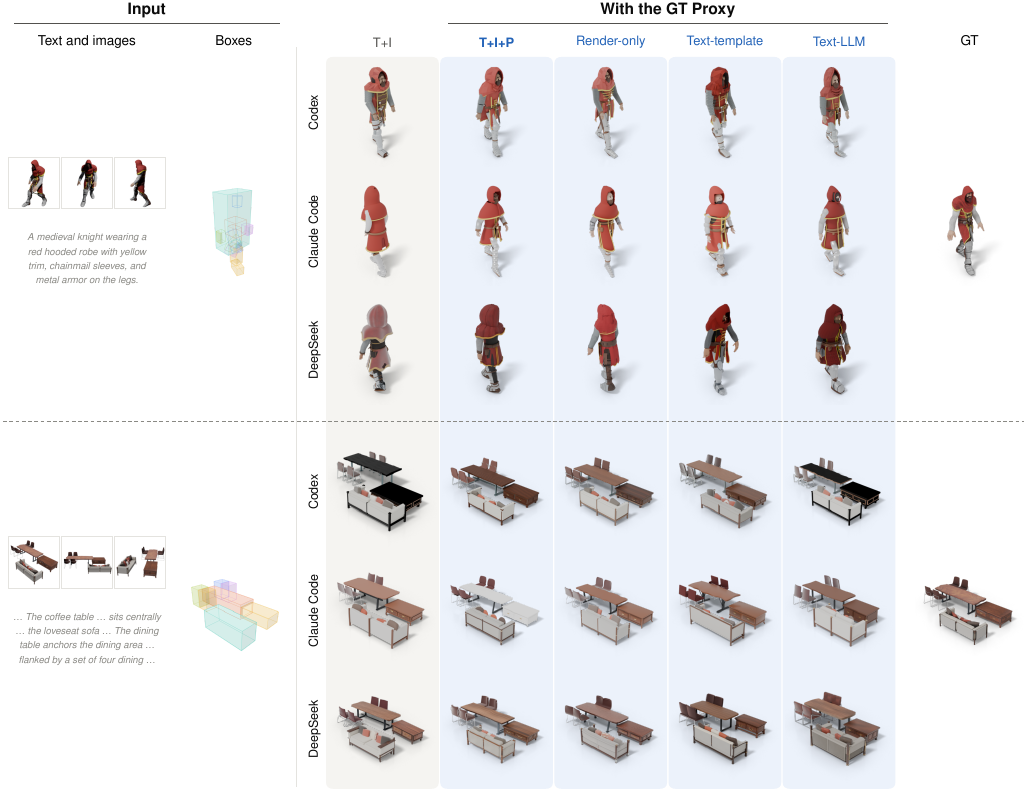}
\caption{\textbf{Models by proxy encoding} for the samples of \cref{fig:encoding_inputs}, one row per agent: the model without a proxy (\sBase, gray), with the GT proxy in each encoding (blue), and the GT. Renders share one rig, each model turned to its best-matching heading. Claude Code's \sGT scene lost its materials at export and renders white.}
\label{fig:encoding_outputs}
\end{figure*}

\mTabEncoding{} averages each proxy encoding over the three agents.
\cref{tab:encoding_all} gives the results of every agent on the same 15-sample subset.
\cref{fig:encoding_inputs} shows the GT proxy in each encoding, and \cref{fig:encoding_outputs}
the models built from them.

\begin{table*}[t]
\centering
\small
\setlength{\tabcolsep}{4pt}
\fittab{\input{tab/tab_encoding_all}}
\caption{\textbf{Proxy encodings, per agent.} As \mTabEncoding{}, for each agent over the same 15 samples. Chamfer distance: geometric mean, $\times 10^{3}$; DINOv3 and Uni3D similarity: mean. \textbf{Bold}: best per agent; \underline{underlined}: second best.}
\label{tab:encoding_all}
\end{table*}

\paragraph{Settings.}
All encodings keep every box's label and, in scenes, its description
(\cref{fig:encoding_inputs}), and the prompt differs from that of \sGT only in naming the file
that holds the proxy. \sRenderOnly is one image of the labeled boxes from a perspective view and
from the front, the right side and the top, on a 0.5\,m grid. \sTextTemplate states, per box, its
size as a fraction of the whole, its position in words (\eg, ``low down, towards the back, right
of centre''), its rotation to the nearest 5$^\circ$ and the boxes it overlaps (137--772 words).
\sTextLLM was written once per sample by Claude Code (Claude Opus 5.5) from all inputs of \sGT,
as a few sentences a user would type, without numbers (111--131 words). Each \sTextLLM text was
checked against the boxes and the reference images, and two sentences of one scene's text were corrected
by hand. The texts are fixed, so every agent reads the same one.

\paragraph{Results.}
\sGT gives the lowest Chamfer distance for every agent, and \sRenderOnly the second lowest for
Codex and Claude Code (\cref{tab:encoding_all}). \cref{fig:encoding_outputs} shows the models.

\section{User Study Details}
\label{sec:supp_user}

This section details the two studies of \mSecUser{}: the online preference
survey, and the modeling study in which participants worked in \uiname.

\subsection{Procedure and Survey Design}
\label{sec:supp_procedure}

\begin{figure}[t]
\centering
\includegraphics[width=\linewidth]{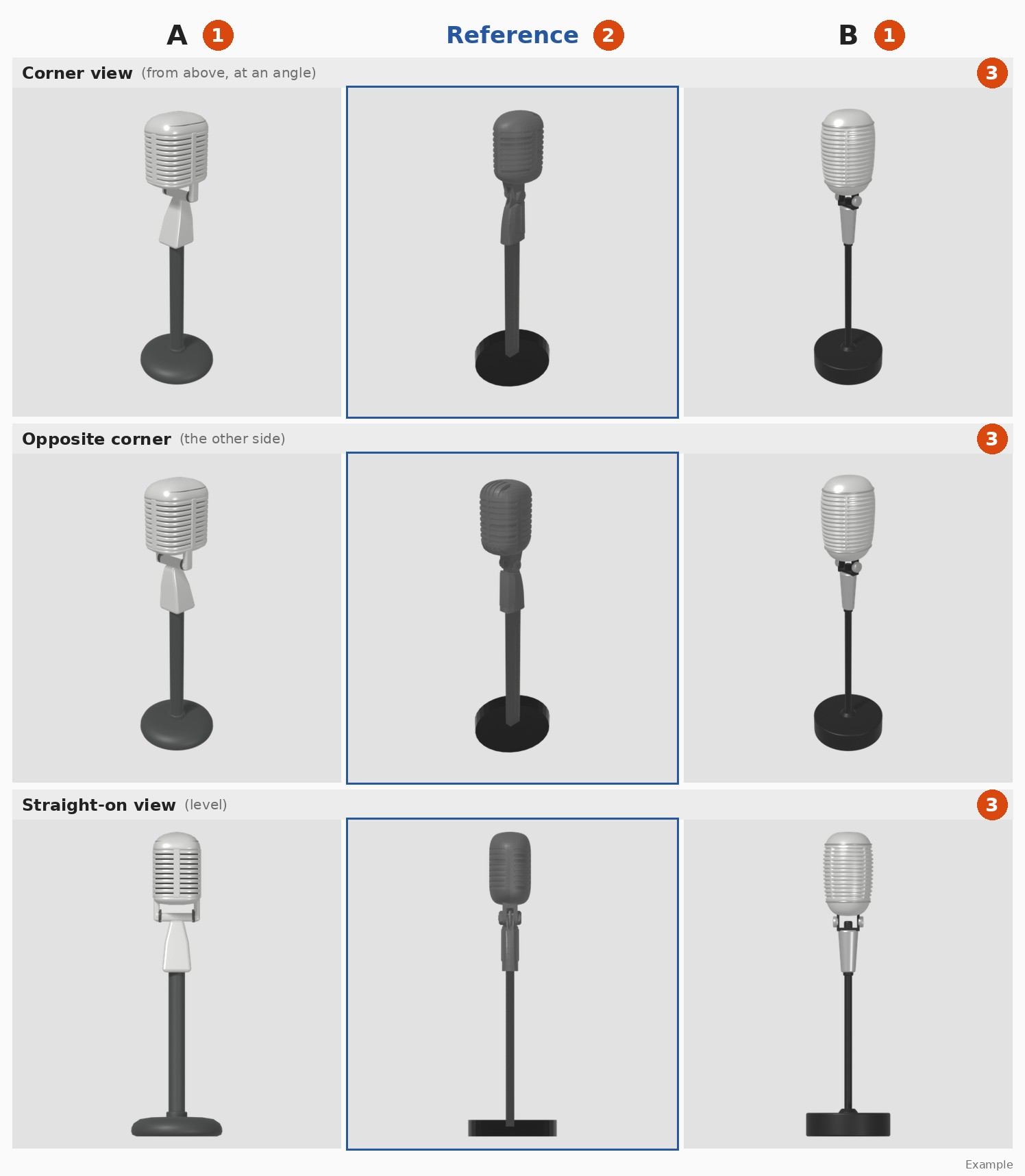}
\caption{\textbf{A question of the preference survey}, as annotated in the survey's instructions: (1) the two models A and B, each built by an agent; (2) the reference; (3) each row shows the three models from the same viewpoint. Above the image, a question shows the caption.}
\label{fig:survey_example}
\end{figure}

\paragraph{Preference survey: design.}
\label{sec:supp_survey}
The survey compares the 120 pairs of models that Codex and Claude Code built for the 60 samples
in \sBase and in \sGT. It ran online. Each participant answers one of ten forms of 28 questions:
one for practice, 24 comparisons, two attention checks and one repeated comparison. The 24
comparisons of a form are 16 pairs of objects and 8 pairs of scenes, half of them from each
agent, with no sample twice. Five forms hold the 120 pairs, and the other five mirror them with
A and B swapped, so every pair is shown with the model built with the proxy once on each side.
The forms are assigned evenly over the participants who start the survey.

\paragraph{Preference survey: questions.}
A question shows the caption and one image (\cref{fig:survey_example}); the long captions of
scenes are folded and open on a click. The image shows the reference between
the two models, from two opposite corner views and one straight-on orthographic view, which looks
from the top for scenes and from the front or the side for objects. The corner views come from
the rig of \cref{sec:supp_metrics}, and all views share one gray background. Each agent's model is
turned to the heading of \cref{eq:turn}, which we checked by eye and corrected for six models,
and each model is framed on its own, so neither the heading nor the scale of a model reveals its
setting. Where a model lost its colors at export, which affects 17 of the 120 pairs, the
question shows the renders that the two agents made of their own models beside the reference
images instead. The proxy is never shown.
Each question asks ``Which model, A or B, better matches the reference (middle) and the
description?'', with the hint ``Example criterion: geometry accuracy, spatial relationship,
visual quality, etc.'', and offers three answers: ``A is closer'', ``No clear difference'' and
``B is closer''. Questions come one per page, and participants cannot go back.

\paragraph{Preference survey: procedure.}
The survey opens with the participant information and a consent box; without consent it ends
there. Participants then state their experience (``How much experience do you have with 3D
modelling or 3D software?'': ``None'', ``Some (occasional use, hobby, coursework)'' or ``A lot
(regular or professional use)'') and read the instructions, which explain the layout of a
question on the annotated example of \cref{fig:survey_example}. The instructions ask for the
closer, more faithful copy of the reference and its description, name geometry accuracy, spatial
relationships and visual quality as example criteria, and tell participants who cannot tell the
two apart to choose ``No clear difference''.
Answers are anonymous, and a complete response took a median of 14\,min.

\paragraph{Preference survey: checks.}
An attention check shows a model against a copy of it with one obvious defect: a major part is
missing, or the model is squashed or stretched. In the first responses the decoy was a model of
another object; these participants found it distracting, and we replaced it. The repeated
comparison shows one of the first 12 comparisons again among the last questions, with A and B
swapped. It checks consistency only and is not counted. We leave out participants who fail both
attention checks, and we do not use unfinished responses.

\paragraph{Preference survey: participants and answers.}
We analyze the 32 valid responses: 15 participants without 3D modeling experience, 15 with some
and two experts. One further participant failed both attention checks and is left out; the 32
passed 63 of their 64 attention checks and answered the repeated comparison the same way in 23
cases. We count the 24 comparisons of each form, so every pair received 4--9 answers. Of the
768 answers, 233 see no clear difference; we filter these out and count the 535 answers that
state a preference. Of these, 361 prefer the model built with the proxy and 174 the model built
from text and images: the proxy wins 67.5\,\% (95\,\% bootstrap interval over samples:
61--74\,\%), and 62\,\%, 71\,\% and 69\,\% on easy objects, hard objects and scenes.
Of the 60 samples, 44 lean to the proxy and 13 to text and
images, and 30 of the 32 participants preferred the proxy's model more
often than the other.

\paragraph{Modeling study: participants and procedure.}
Seven PhD students took part: two with no 3D modeling experience, two with some Blender
experience, two with some CAD experience, and one with extensive CAD experience. Sessions were
held in person, one participant at a time, with the experimenter present, on one CPU-only
machine. Each session followed the same steps:
(1) participant information and consent;
(2) an introduction to the project and to the motivation for the proxy, shown with
\mFigIdeation{};
(3) a one-page cheat sheet of the controls of \uiname (\cref{sec:supp_ui}) and a demonstration
by the experimenter;
(4) time for questions and free practice;
(5) the three tasks, recorded under the participant's ID, each followed by two short questions;
(6) the final questions; and
(7) an optional free modeling session, which we do not analyze.
During the tasks, \uiname logs every interaction with its time: each action on a box, each edit
of the text and the images, and each launch, message and interrupt.

\paragraph{Tasks.}
As given to participants:
\begin{itemize}
\item \emph{T1.} Choose one object from the easy category. Describe it with text, images and
labeled 3D bounding boxes in the studio, and launch the agent to model it.
\item \emph{T2.} Choose one object from the hard category. Describe it with text and images
only, and launch the agent to draft the labeled 3D bounding boxes. Review the boxes, modify
them or ask the agent to modify them further if needed, then launch the agent to model the
object.
\item \emph{T3.} Choose one scene from the rooms category. Think of one modification you would
like to make to the scene, \eg, add, move, resize or rearrange furniture. Describe it with
any method(s), including text, images and labeled 3D bounding boxes, and launch the agent to
edit the scene accordingly.
\end{itemize}
Participants choose their targets in a gallery that shows the test sets as 3D views, without
captions: the GT meshes of the easy objects for T1 and of the hard objects for T2, and for T3 the
20 scenes as Codex modeled them in \sGT. In T3, the participant continues the agent session that
built the chosen scene. Participants provide their own reference images: for an object, one to
seven screenshots of the target in the gallery. One participant added a photo from the web for
the colors of a target that has none.

\paragraph{Modeling study: questions.}
The experimenter asked all questions orally and noted the answers; the questions are open, with
no rating scale. After each task, participants were asked how well the final model matched what
they intended, and one question specific to the task: how much the boxes helped (T1), how good
the agent-drafted boxes were (T2), and why they chose the input methods they used (T3). At the
end:
\begin{itemize}
\item Over all tasks, which input method did you rely on the most, and which the least, and
why?
\item Was there anything you wanted to describe but felt none of the methods could express
well enough?
\item How much did you feel in control of the final results?
\item (For experienced 3D modelers) Compared with your usual modeling approach, where does
this method save effort, and where does it cost more?
\item If you could change or add one thing in this system, what would it be?
\end{itemize}

\subsection{Interaction Traces}
\label{sec:supp_traces}

\begin{figure*}[t]
\centering
\includegraphics[width=\textwidth]{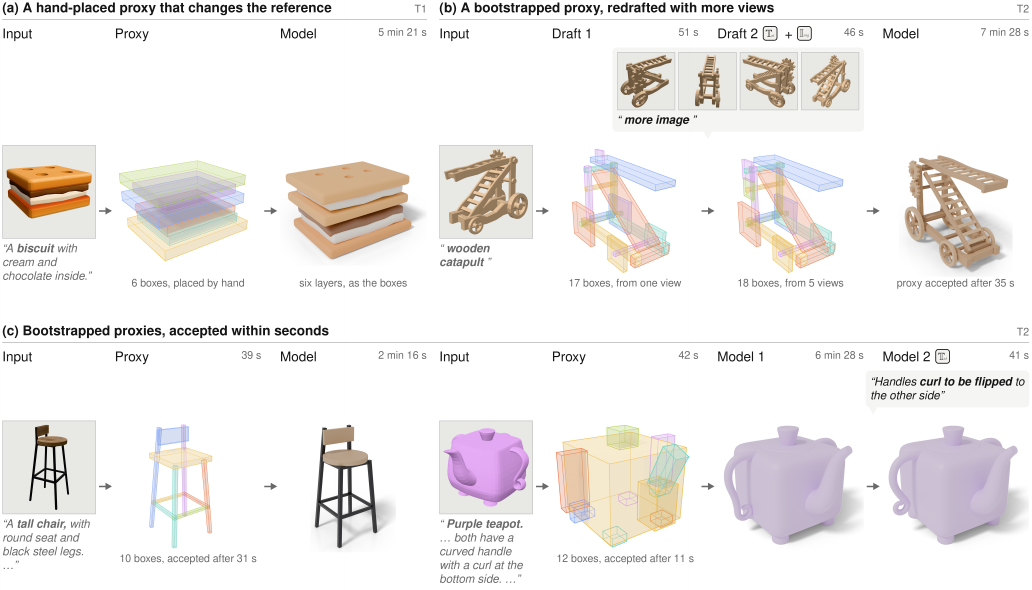}
\caption{\textbf{Interaction traces from the modeling study.} Four sessions, turn by turn, with Claude Code as the agent. Each message is shown as typed, over the version that it led to, with the agent's run time. (a) T1: the reference image shows a biscuit of four layers. The participant stacks six boxes, and the model has six layers, as the boxes. (b) T2: the agent bootstraps a first proxy from one view. The participant adds the four views shown with the message, and the agent redrafts. The participant accepts the second proxy 35\,s after it is drafted, and the agent models from it. (c) T2: two participants accept the bootstrapped proxy as drafted, 31\,s and 11\,s after the agent finishes it. For the teapot, one message then flips the curl of the handles. Two versions were overwritten by later turns and are recovered from the session's files: draft~1 of (b) from the agent's command, and model~1 of the teapot from Blender's backup of the scene.}
\label{fig:user_traces}
\end{figure*}

We show six sessions of the study turn by turn, each for one observation.
\cref{fig:user_traces} has four: (a) a proxy placed by hand that changes the reference, where the
model takes its structure from the boxes; (b) a bootstrapped proxy that the participant has
redrafted after adding views; and (c) two bootstrapped proxies that their participants accept
within seconds. \mFigTrace{} follows an edit of a room with text and boxes, and
\cref{fig:failure_curvature} a bootstrapped proxy that the participant corrects in text.

\begin{table}[t]
\centering
\small
\setlength{\tabcolsep}{4pt}
\begin{tabular}{lccc}
\toprule
 & T1 & T2 & T3 \\
\midrule
Task segment (min) & 12.3 & 12.7 & 15.4 \\
Start to first launch or message (min) & 5.6 & 2.6 & 3.8 \\
Input activity (min) & 7.8 & 5.9 & 5.5 \\
Agent runtime (min) & 5.4 & 7.9 & 4.3 \\
Boxes at modeling launch & 6 & 18 & -- \\
\bottomrule
\end{tabular}
\caption{\textbf{Interaction statistics}, median over the seven participants. A task segment includes inspection, conversation and idle time, so it is not a completion time. Input activity sums the gaps of at most 30\,s between input events; it overlaps agent runtime, so the two do not add up. Agent runtime counts every run started in the task, including runs that continue into the next one.}
\label{tab:study_stats}
\end{table}

\paragraph{Recordings.}
The studio logged every interaction (\cref{tab:study_stats}). Agent runtime (including Blender rendering) is substantial on
our CPU-only study machine. In T2, nobody added, removed, moved or edited a drafted box by hand; two
participants also selected boxes during review. Participants interrupted five runs since the partial result already satisfied them.
In one T3 session, a failed send dropped the note that tells the agent which inputs
changed, so the agent edited the room from the text alone and did not read the participant's boxes.

\begin{figure*}[t]
\centering
\includegraphics[width=\textwidth]{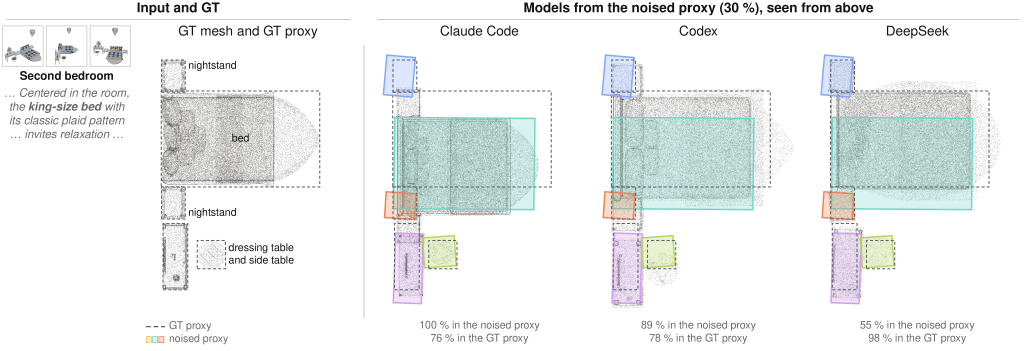}
\caption{\textbf{Conflicting inputs.} A bedroom modeled from a proxy at 30\,\% noise, seen from above. Left: the reference images, the caption (abridged) and the GT mesh with the GT proxy. Right: each agent's model (points) with the noised proxy that it was given (color) and the GT proxy (dashed). The noised proxy moves the bed onto one of its nightstands, while the reference images show it between the two. Claude Code builds the bed in its noised box, DeepSeek builds it where the images show it, and Codex stays close to the noised boxes. Below each model: the share of its surface inside the noised proxy and inside the GT proxy.}
\label{fig:failure_conflict}
\end{figure*}

\paragraph{Results.}
Each participant replicated an easy object with text, images and a proxy placed by hand (T1),
replicated a hard object with a proxy that the agent drafted and they reviewed (T2), and edited
an agent's model of a room by any means (T3). All 21 tasks ended with a model. Claude Code (Claude Opus
5.5) modeled in T1 and T2, and Codex (GPT-6.1 Sol) continued its own room in T3, both at medium
effort. The three tasks took 29--61\,min per participant. In T2, five participants accepted the
first draft (\cref{fig:user_traces}c), one asked for a redraft after adding views
(\cref{fig:user_traces}b), and one asked the agent in text, twice, to
correct the boxes; they accepted the draft a median of 31\,s after the agent finished it. The
drafts held 10--30 boxes, each labeled and described in detail, against 2--8 boxes placed by hand
in T1. Five participants named their boxes and two left them unnamed; nobody wrote a description
for a box.

\paragraph{Feedback.}
With hand-placed boxes (T1), participants found that the models followed the proxy's positions
and proportions (\cref{fig:user_traces}a). Where the proxy and the images disagreed, the agent weighed the two and
reported where it departed from the boxes: it kept one participant's box sizes against the image,
placed a seat as in the image when its box sat behind the rear wheel, and evened out legs whose
boxes were unevenly placed. Two participants started T1 with text and images alone and added
boxes after a reminder.
In T2, all were satisfied with the resulting model; one noted that it captured the main shapes
and saved them time. Two asked for one correction to the direction of a curved part (a teapot's
handles, a chair's back; \cref{fig:user_traces}c, \cref{fig:failure_curvature}), which boxes do not express, and one found
such curvature hard to describe with any of the inputs. Over the 14 object tasks of T1 and T2, four results followed one
corrective message.
For the open-ended edits of T3, all seven wrote text, three also placed boxes and two added
images. Participants described coarse or semantic changes in text and placed boxes where the
exact position, size or orientation mattered (\mFigTrace{}). Those who wrote text
alone chose it as the quickest input, and one of them found an added object away from where they
had imagined it. In two edits the agent also made a choice that the participant had not asked
for (a laptop on a new desk, the gender of a sleeping person).
They described a division of labor between the inputs: images for what things look like, the
proxy for where things go and how large they are, and text for everything else. The input they
relied on most varied: images, boxes and text were each named, and one participant said it
depended on the task. Two participants relied on boxes least, as text and images were familiar
to them from other tools, and three wrote in text what their images already showed.
Six of the seven felt in control of the results; one wanted more back-and-forth with the agent
before judging, and one tied their control to the amount of input they gave.
Participants saw the tool as suited to prototyping and creative work rather than precise
mechanical design. Requested features include agents that adjust the user's boxes while
modeling, boxes created automatically for newly added objects or from 2D sketches, checks of
units and scale, further views of a reference generated from a single image, and grouping of
boxes.

\subsection{Preference and Automated Metrics}
\label{sec:supp_correlation}

Per pair, we compare the net preference (answers for the proxy minus answers against it, over
all answers) with the gain that the proxy brings in each metric of \mTabMain{}.
The preference correlates with all three: Spearman's $\rho$ is 0.23 for Chamfer distance (as
the log ratio), 0.36 for DINOv3 and 0.33 for Uni3D, while the
gains in the three metrics correlate only weakly with each other (0.11--0.37).
Participants weigh both shape and appearance, and each metric captures part of what they judge.
Grouped into thirds by the proxy's Chamfer gain, the pairs give the proxy 59\,\%, 70\,\% and
73\,\% of the answers that state a preference.

\section{More Discussions}
\label{sec:supp_discussion}

The proxy adds spatial control to text and images, and it has limits of its own. We discuss
three typical failure cases and the future work that they point to.

\begin{figure*}[t]
\centering
\includegraphics[width=\textwidth]{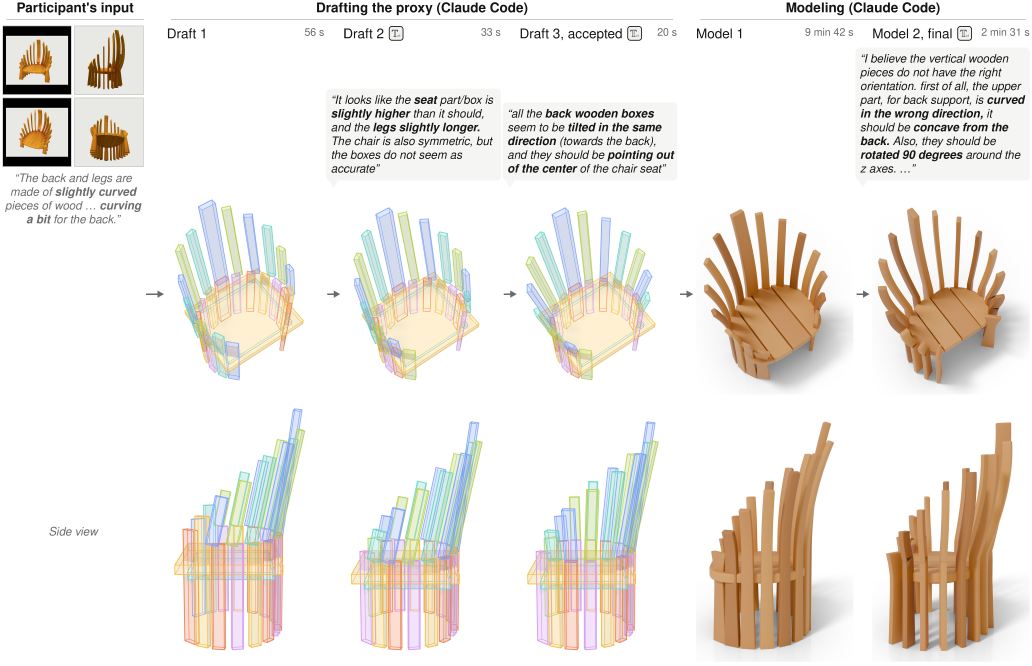}
\caption{\textbf{Curved parts.} A participant of the modeling study models a chair of curved ribs (task T2). Left: the reference images and the text they gave (abridged). Middle: the proxy that Claude Code bootstrapped, and its two revisions on the participant's messages; every rib is one straight box. Right: the first model, whose back ribs bend the wrong way, and the final model after one more message. Each message is shown above the version that it led to, with the agent's run time. Top row: three-quarter view; bottom row: side view.}
\label{fig:failure_curvature}
\end{figure*}

\paragraph{Conflicting inputs.}
\label{sec:supp_failures}
A new kind of input brings a new requirement: the text, the images and the proxy have to agree
with one another. A proxy is rarely exact, whether a user places it or an agent bootstraps it, so
its boxes sometimes contradict what the agent sees in the reference images. Our prompt calls the
boxes approximate references and asks for the joint expectation of the three inputs
(\cref{sec:supp_prompts}), which leaves the agent to decide which input to follow.

The noised proxies of our experiments show such conflicts. At 10\,\% noise, the level of \sNoisy,
the proxy and the images still agree, and the models keep most of the gain of the GT proxy
(\cref{fig:supp_geometry}). At 30\,\%, the boxes move far enough to contradict the images, and the
agents respond differently. In the bedroom of \cref{fig:failure_conflict}, the noised proxy shifts
the bed by about half a meter, onto one of its two nightstands, while the reference images show
the bed between them. Claude Code builds the bed in the shifted box, moves the nightstands to
either side of it and reports the change. Codex follows the noised boxes without comment. DeepSeek
builds the bed where the images show it. As a result, all of Claude Code's model lies inside the
noised proxy and 76\,\% of it inside the GT proxy, against 55\,\% and 98\,\% for DeepSeek. Each
agent settles the same conflict in favor of a different input, so a user with a rough proxy
cannot tell in advance which input will win. A participant of our modeling study met the same
conflict with boxes placed by hand: the agent followed the boxes, and one message turned it to the
image (\cref{sec:supp_traces}). In \uiname, the model appears
over the boxes when the agent finishes, so the user sees where the two differ and corrects the
boxes or the model in the next turn.

\paragraph{Texture and visual style.}
\label{sec:supp_appearance}
The proxy is a spatial input. It tells the agent where each part goes, how large it is and what
it is called, and it says nothing about texture, material or visual style. The look of a model
still comes from the text and the reference images, so a model can follow its proxy closely and
miss the texture or the style that the user had in mind. The participants of our modeling study
worked accordingly. They described a division of labor between the inputs, with images for what
things look like and the proxy for where things go and how large they are, and one of them added
a photo from the web to give a target its colors (\cref{sec:supp_procedure,sec:supp_traces}).
The preference survey shows why this matters. Participants weigh appearance as well as shape,
and their preference follows the proxy's gain in the DINOv3 similarity of the renders more
closely than its gain in Chamfer distance (\cref{sec:supp_correlation}).

\paragraph{Curved shapes.}
\label{sec:supp_limits}
The shape inside a box is open as well: a box states where a part is, how large it is and which
way it is turned. Boxes therefore have no simple way to describe a curved part. One box bounds the
curve and leaves its direction open. A chain of small boxes can follow the curve piece by piece,
but such a proxy is tedious to place and to edit, which defeats its purpose. For now, users
describe the exact curve in the text or show it in the images. \cref{fig:failure_curvature}
follows a participant of our modeling study through a chair made of curved ribs. The bootstrapped
proxy holds every rib in one straight box, and the participant revises it in two messages, the
second about the direction in which the back ribs tilt. The first model then bends the back ribs
the wrong way, although the text asks for a curved back and the images show it, and a third
message corrects it. Two of the seven
participants asked for such a correction of a curved part, and one found curvature hard to
describe with any of the inputs (\cref{sec:supp_traces}).

\paragraph{Future work.}
The last two cases point to inputs that say more than a box does. For curved shapes, the next
input is the hand-drawn sketch. With a sketch tool in \uiname, a user draws any curve they want
directly on the model, in 3D, and the strokes enter the workspace next to the boxes;
participants of the modeling study asked for sketch input as well (\cref{sec:supp_traces}). For
texture and style, a reference image could be bound to a box, so that the box states where a
part goes and the image shows how it looks. More generally, richer proxies that stay 3D-native
and agent-native, such as other primitives, 3D sketch curve networks or coarse meshes, could
carry shape as well as layout.

%% file: tab/tab_all_metrics_mean.tex
\begin{tabular}{@{}lrrrrrrrr@{}}
\toprule
 & \multicolumn{3}{c}{Geometry vs.\ GT} & \multicolumn{2}{c}{Appearance vs.\ GT} & \multicolumn{3}{c}{Agreement with inputs} \\
\cmidrule(lr){2-4}\cmidrule(lr){5-6}\cmidrule(lr){7-9}
Setting & \multicolumn{1}{c}{\begin{tabular}{@{}c@{}}Chamfer\\($\times 10^{3}$) $\downarrow$\end{tabular}} & \multicolumn{1}{c}{\begin{tabular}{@{}c@{}}F-score\\@0.05 $\uparrow$\end{tabular}} & \multicolumn{1}{c}{\begin{tabular}{@{}c@{}}sv-IoU $\uparrow$\end{tabular}} & \multicolumn{1}{c}{\begin{tabular}{@{}c@{}}DINOv3\\render–render $\uparrow$\end{tabular}} & \multicolumn{1}{c}{\begin{tabular}{@{}c@{}}Uni3D\\shape–shape $\uparrow$\end{tabular}} & \multicolumn{1}{c}{\begin{tabular}{@{}c@{}}SigLIP 2\\text–render $\uparrow$\end{tabular}} & \multicolumn{1}{c}{\begin{tabular}{@{}c@{}}SigLIP 2\\text–agent render $\uparrow$\end{tabular}} & \multicolumn{1}{c}{\begin{tabular}{@{}c@{}}DINOv3 agent\\render–ref. $\uparrow$\end{tabular}} \\
\midrule
\textit{\sBaseFull} & 6.9 & 0.66 & 0.16 & 0.824 & 0.823 & \textbf{0.145} & \underline{0.141} & 0.793 \\
\textbf{\textit{\sGTFull}} & \textbf{3.5} & \textbf{0.77} & \textbf{0.22} & \textbf{0.840} & \textbf{0.835} & \textbf{0.145} & \underline{0.141} & \textbf{0.804} \\
\textit{\sNoisyFull} & \underline{4.0} & \underline{0.75} & \underline{0.20} & \underline{0.831} & \underline{0.830} & \textbf{0.145} & \underline{0.141} & \underline{0.802} \\
\textit{\sDraftedFull} & 6.1 & 0.68 & 0.16 & 0.820 & 0.823 & 0.144 & \textbf{0.142} & 0.796 \\
\bottomrule
\end{tabular}

%% file: tab/tab_encoding_all.tex
\begin{tabular}{@{}lrrrrrrrrr@{}}
\toprule
 & \multicolumn{3}{c}{Codex} & \multicolumn{3}{c}{Claude Code} & \multicolumn{3}{c}{DeepSeek} \\
\cmidrule(lr){2-4}\cmidrule(lr){5-7}\cmidrule(lr){8-10}
Setting & \multicolumn{1}{c}{Chamfer$\downarrow$} & \multicolumn{1}{c}{DINOv3$\uparrow$} & \multicolumn{1}{c}{Uni3D$\uparrow$} & \multicolumn{1}{c}{Chamfer$\downarrow$} & \multicolumn{1}{c}{DINOv3$\uparrow$} & \multicolumn{1}{c}{Uni3D$\uparrow$} & \multicolumn{1}{c}{Chamfer$\downarrow$} & \multicolumn{1}{c}{DINOv3$\uparrow$} & \multicolumn{1}{c}{Uni3D$\uparrow$} \\
\midrule
\textit{\sBaseFull} & 7.0 & \underline{0.849} & 0.840 & 6.6 & 0.816 & \textbf{0.849} & \underline{7.5} & 0.798 & 0.812 \\
\textbf{\textit{\sGTFull}} & \textbf{3.0} & \textbf{0.856} & \textbf{0.868} & \textbf{2.9} & \textbf{0.835} & 0.844 & \textbf{4.0} & 0.807 & 0.800 \\
\midrule
\textit{\sRenderOnly} & \underline{4.2} & 0.835 & 0.846 & \underline{3.2} & \underline{0.827} & \underline{0.845} & 7.8 & \textbf{0.813} & 0.816 \\
\textit{\sTextTemplate} & 5.4 & 0.837 & 0.851 & 4.0 & 0.811 & 0.815 & 8.4 & 0.795 & \textbf{0.822} \\
\textit{\sTextLLM} & 4.6 & 0.827 & \underline{0.857} & 5.8 & 0.800 & 0.829 & 7.7 & \underline{0.810} & \underline{0.818} \\
\bottomrule
\end{tabular}